\documentclass{article}

\usepackage[preprint]{neurips_2026}

\usepackage[utf8]{inputenc} %
\usepackage[T1]{fontenc}    %
\usepackage{hyperref}       %
\usepackage{url}            %
\usepackage{booktabs}       %
\usepackage{amsfonts}       %
\usepackage{nicefrac}       %
\usepackage{microtype}      %

\title{Unlocking the Working Memory of \\Large Language Models for Latent Reasoning}

\usepackage[most]{tcolorbox}
\usepackage{inconsolata}
\usepackage{multirow}
\usepackage[table]{xcolor}

\newcommand{\method}{RiM}
\newcommand{\fullmethod}{Reasoning in Memory}
\usepackage{pdfrender}

\DeclareRobustCommand{\best}[1]{\textbf{#1}}

\definecolor{OursMainYellow}{HTML}{FDF6EE}
\definecolor{OursSoftYellow}{HTML}{FEFBF7}

\usepackage{wrapfig}

\usepackage{rotating}
\usepackage{subcaption}
\usepackage{graphicx}

\usepackage{arydshln}
\definecolor{C2}{RGB}{68, 134, 83}

\usepackage{algorithm}
\usepackage[noend]{algorithmic}

\usepackage{color}
\usepackage{amsmath}
\usepackage{amssymb}
\usepackage{amsthm}
\usepackage{mathtools}
\usepackage{thmtools}
\usepackage[mathscr]{eucal}
\usepackage{bm}
\usepackage{bbm}
\usepackage{relsize}
\usepackage{graphics}
\usepackage{wrapfig}
\usepackage{multirow}
\usepackage{enumitem} 

\usepackage[capitalize,noabbrev]{cleveref}

\usepackage{pgf}
\usepackage{xinttools}
\usepackage{tikz}
\usetikzlibrary{arrows.meta}
\usepackage{textcomp}
\usetikzlibrary{calc,shapes,arrows,positioning,automata,trees}
\usepackage{placeins} 
\usepackage{tabularx}
\usepackage{graphicx}
\usepackage{subcaption} 
\usepackage{listings}
\usepackage{xargs,lipsum,caption,changepage,ifthen}

\usepackage{tikz} 

\usepackage{footmisc}

\usepackage{bigdelim}
\usepackage{colortbl} 
\definecolor{mColor1}{rgb}{0.95,0.95,0.95}
\newcolumntype{a}{>{\columncolor{mColor1}}c}

\usepackage[toc,page]{appendix}

\usepackage{tcolorbox}

\definecolor{solarized@base03}{HTML}{002B36}
\definecolor{solarized@base02}{HTML}{073642}
\definecolor{solarized@base01}{HTML}{586e75}
\definecolor{solarized@base00}{HTML}{657b83}
\definecolor{solarized@base0}{HTML}{839496}
\definecolor{solarized@base1}{HTML}{93a1a1}
\definecolor{solarized@base2}{HTML}{EEE8D5}
\definecolor{solarized@base3}{HTML}{FDF6E3}
\definecolor{solarized@yellow}{HTML}{B58900}
\definecolor{solarized@orange}{HTML}{CB4B16}
\definecolor{solarized@red}{HTML}{DC322F}
\definecolor{solarized@magenta}{HTML}{D33682}
\definecolor{solarized@blue}{HTML}{268BD2}
\definecolor{solarized@cyan}{HTML}{2AA198}
\definecolor{solarized@green}{HTML}{859900}

\newtcolorbox{importantresult}{colback=solarized@yellow!5!white,
colframe=solarized@yellow,parbox, left=0.5mm, right=0.5mm,top=0.5mm,bottom=0.5mm}

\newtcolorbox{importantresult_noparbox}{colback=solarized@yellow!5!white,
colframe=solarized@yellow,parbox=false, left=0.5mm, right=0.5mm,top=0.5mm,bottom=0.5mm}

\newtheorem*{theorem*}{Theorem}
\newtheorem*{definition*}{Definition}

\newcommand\Bm{\bm{m}}

\newcommand\Br{\bm{r}}

\newcommand\Bw{\bm{w}}
\newcommand\Bx{\bm{x}}
\newcommand\By{\bm{y}}
\newcommand\Bz{\bm{z}}

\DeclarePairedDelimiterX{\kldiv}[2]{(}{)}{%
  #1\,\delimsize\|\,#2%
}

\DeclarePairedDelimiterX{\mi}[2]{(}{)}{%
  #1\;\delimsize ; \;#2%
}

\DeclarePairedDelimiterX{\di}[2]{(}{)}{%
  #1\;\delimsize ; \;#2%
}

\DeclarePairedDelimiterX{\ce}[2]{(}{)}{%
  #1  \mathbin{;} \, #2%
}

\DeclarePairedDelimiterXPP{\mii}[3]%
   {_{\mathrm{#1}}}{(}{)}{}{#2\;\delimsize ; \;#3%
}

\makeatletter
\newcommand{\dlmf}[1]{%
\citep[%
  \def\nextitem{\def\nextitem{, }}%
  \@for \el:=#1\do{\nextitem\href{http://dlmf.nist.gov/\el}{(\el)}}%
]{Olver:10}%
}
\makeatother

\usepackage{pdflscape} 

\newcolumntype{R}[1]{>{\raggedright\arraybackslash}p{#1}}
\newcolumntype{C}[1]{>{\centering\arraybackslash}p{#1}}
\newcolumntype{L}[1]{>{\raggedleft\arraybackslash}p{#1}}

\usepackage{bigdelim}
\usepackage{colortbl} 
\definecolor{mColor1}{rgb}{0.95,0.95,0.95}

\hypersetup{
   colorlinks=true,
   linkcolor=[RGB]{30, 30, 180},
   citecolor=[RGB]{30, 30, 180},
   urlcolor=magenta,
   pdfborder=0 0 0,
   pdftitle={},
   pdfsubject={}, 
   pdfkeywords={},
   pdfauthor={},%
   pdfstartview=FitH
}

\renewcommand{\paragraph}[1]{\textbf{#1}}

\usepackage[T1]{fontenc}
\usepackage{inconsolata}

\usepackage{makecell} %
\usepackage{colortbl}
\usepackage{arydshln}
\definecolor{G}{RGB}{0, 0, 0}
\colorlet{G_4}{G!4!white}
\newcolumntype{g}{>{\columncolor{G_4}}l}
\newcolumntype{G}{>{\columncolor{G_4}}c}
\colorlet{G_2}{G!2!white}
\newcolumntype{A}{>{\columncolor{G_2}}c}

\definecolor{citeblue}{HTML}{48869D}
\hypersetup{
  colorlinks=true,
  urlcolor=citeblue,
  linkcolor=citeblue,
  citecolor=citeblue
}

\author{%
}
\author{
  Lukas Aichberger $^{1}$ \ \
  Sepp Hochreiter $^{1,2}$ \\ \\
$^1$~ELLIS Unit Linz and LIT AI Lab, Institute for Machine Learning, \\ 
Johannes Kepler University Linz, Austria \\
$^2$~NXAI GmbH, Linz, Austria \\
\texttt{\{aichberger, hochreit\}@ml.jku.at}
}

\begin{document}

\maketitle

\begin{abstract}

To improve the reasoning capabilities of large language models, test-time compute is typically scaled by generating intermediate tokens before the final answer.
However, this couples reasoning to autoregressive generation and thereby conflates internal computation with external communication.
In contrast, human cognition can use working memory to hold and manipulate information internally without the need to externalize intermediate thoughts.
Drawing on this principle, we introduce \emph{\fullmethod{} (\method{})}, a latent reasoning method that replaces the autoregressive generation of reasoning steps with memory blocks.
These memory blocks are fixed sequences of special tokens that unlock the working-memory capacity of large language models. Since they are fixed rather than generated, they can be processed in a single forward pass, enabling compute-efficient latent reasoning.
To operationalize these memory blocks, we employ a two-stage curriculum. First, we ground them by predicting explicit reasoning steps after each memory block. Second, we discard this step-level supervision and iteratively refine the final answer after each memory block.
Our experiments on reasoning benchmarks show that, across language models of different families and sizes, \method{} matches or exceeds existing latent reasoning methods while avoiding the autoregressive generation of thoughts. These results demonstrate that large language models can be trained to use working memory as an effective mechanism for latent reasoning.

\end{abstract}

\section{Introduction} \label{sec:introduction}

\begin{figure}
    \centering
    \includegraphics[width=\linewidth, trim = 0.0cm 0.0cm 0.0cm 0.0cm, clip]{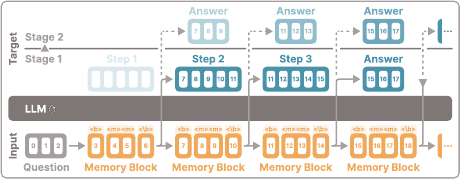}
    \caption{\textbf{Reasoning in Memory (\method{}).}
    Stage~1 trains the LLM to use memory blocks (yellow) as working memory by supervising the prediction of the next reasoning step (blue) after each memory block. Once the memory blocks are grounded for intermediate computation, Stage~2 removes reasoning-step supervision and trains the LLM to refine the final answer after each memory block.}
    \label{fig:figure_1}
    \vspace{-2pt}
\end{figure}

Large language models (LLMs) have demonstrated remarkable reasoning capabilities, primarily driven by techniques that scale test-time compute through generating intermediate tokens before giving the final answer \citep{Wei:22, Snell:24}.
In chain-of-thought (CoT) reasoning, intermediate computation remains directly coupled to the generation of text. This coupling forces an LLM to ``think out loud'', anchoring its reasoning process to the syntax and structure of natural language. However, language is optimized for communication rather than for computation \citep{Wei:22, Kojima:22}. Consequently, part of the computational budget is allocated for generating grammatical and fluent intermediate text, rather than being devoted purely to internal computation.

Recent advances in latent reasoning bypass these natural-language constraints by replacing discrete tokens with continuous representations \citep{Hao:25, Shen:25, Cheng:24}. 
However, they preserve the same step-by-step generation paradigm, effectively ``thinking out loud'' in a continuous space. 
Whether generating discrete tokens or continuous representations, each intermediate computation must still be externalized before future computation can condition on it. 
Thus, these latent reasoning methods preserve the coupling of intermediate computation to autoregressive generation.

Human cognition suggests a different design principle. When solving complex problems, humans do not typically articulate every intermediate step, nor is their reasoning strictly bound to the syntax of language. Instead, cognitive psychology identifies \emph{working memory} as an internal workspace for holding and manipulating task-relevant representations \citep{Baddeley:92}. Developmental psychology observes that while we initially rely on ``thinking out loud'', this linguistic scaffolding is eventually abandoned in favor of internal processing \citep{Vygotsky:78}.
This motivates a shift from generating intermediate reasoning steps autoregressively to representing them in a dedicated internal workspace of an LLM, analogous to human working memory.

Drawing on these cognitive principles, we introduce \emph{\fullmethod{} (\method{})}, which operationalizes \emph{memory blocks}, fixed sequences of special tokens, as a working memory for latent reasoning.
Rather than generating intermediate reasoning steps autoregressively, our method trains the LLM to use these memory blocks as an internal latent workspace for task-relevant computation.
While their token identities and positions are fixed, their contextual representations remain task-dependent and can encode intermediate reasoning.
A direct benefit of these fixed memory blocks is that they can be processed in a single forward pass.
This avoids the bottleneck of autoregressively generating discrete tokens or continuous representations, making our method highly parallelizable and compute-efficient. 

\begin{wrapfigure}{r}{0.47\textwidth}
    \vspace{-13.8pt}
    \centering
    \includegraphics[width=\linewidth, trim=0.0cm 0.0cm 0.0cm 0.0cm, clip]{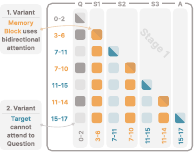}
    \caption{\textbf{\method{} Attention Mask.} Memory blocks (yellow) attend to the question and previous memory blocks. Target reasoning steps (blue) attend to previous memory blocks and optionally the question, but not to other reasoning steps. This enables all targets to be predicted in one forward pass without information leakage, forcing reasoning inside the memory blocks.}
    \vspace{-16pt}
    \label{fig:rim_attention_mask}
\end{wrapfigure}

In order to teach the LLM to use these memory blocks as a latent workspace, \method{} employs a two-stage curriculum (\Cref{fig:figure_1}). 
Since the memory blocks have no predefined computational role, unlocking their working-memory capacity requires a carefully structured training signal. In Stage~1, we therefore ground the memory blocks by supervising the readout after each block to predict the corresponding next reasoning step. As each readout can recover this target only from memory blocks seen so far, the LLM is forced to structure the latent workspace around task-relevant intermediate information.
In Stage~2, once the working-memory capacity is learned, we remove this strict step-level supervision and instead supervise the readout after each block to predict the final answer. The training signal is therefore shifted from predicting written reasoning steps to progressively refining the answer across memory blocks. A custom attention mask (\Cref{fig:rim_attention_mask}) allows all readouts to be trained in a single forward pass, making supervision dense and efficient in both stages (see \Cref{appendix:sec:training_details} for details). %

Empirically, we validate \method{} by training language models across GPT-2 and Llama-3.2 scales on GSM8K-Aug and evaluating them on GSM8K and GSM-Hard as in- and out-of-distribution benchmarks, respectively.
Across these settings, \method{} matches or exceeds existing CoT and latent reasoning baselines.
These findings show that our method enables LLMs to use working memory for efficient reasoning by combining dense supervision during training with compute-efficient inference.

Our main contributions are as follows:
\begin{itemize}[leftmargin=*, topsep=0pt, itemsep=6pt, partopsep=0pt, parsep=0pt]
    \item We introduce \fullmethod{} (\method{}), a latent reasoning method that unlocks the working-memory capacity of LLMs by providing a latent workspace for intermediate computation.
    \item We propose a two-stage curriculum that assigns memory blocks a computational role through carefully structured training signals, enabling their use as working memory for latent reasoning.
    \item We show that \method{} matches or exceeds explicit and latent reasoning baselines across model scales and reasoning benchmarks, while making intermediate computation parallelizable and efficient.
\end{itemize}

\section{Related Work} \label{sec:related_work}

\paragraph{CoT reasoning.}
The modern test-time-compute paradigm grew out of methods that place a textual workspace between the question and the answer \citep{Nye:21}. 
Chain-of-thought (CoT) prompting \citep{Wei:22,Kojima:22} showed that language models solve harder tasks when intermediate computation is externalized as text. 
First, explicit reasoning has been scaled at inference time by repeated sampling \citep{Brown:24}, aggregating multiple reasoning traces \citep{WangSC:23}, structured search \citep{Yao:23}, or process reward models that verify intermediate steps \citep{Lightman:24,Snell:24}. 
Second, it has also been scaled at training time through supervised fine-tuning on self-generated \citep{Zelikman:22} or ground-truth \citep{Muennighoff:25} reasoning traces, or through reinforcement learning \citep{DeepSeek:25}. 
These methods share that intermediate computation is externalized by autoregressively generating text.

\paragraph{Explicit latent reasoning.}
To avoid the syntactic overhead of explicit reasoning, recent work explores latent reasoning.
One line of work performs vertical latent reasoning by iteratively refining activations through recurrent modules \citep{WangHRM:25,Martineau:25}.
A second line of work performs horizontal latent reasoning by replacing parts of the reasoning trace with continuous representations along the sequence dimension.
Coconut \citep{Hao:25} is the closest latent analogue of CoT, as it replaces discrete reasoning tokens with continuous representations that are fed back autoregressively.
Related methods modify how these continuous representations are trained or used, for example through compressed contemplation tokens \citep{Cheng:24}, distillation from explicit CoT \citep{Shen:25}, synthetic continuous targets \citep{WangSynAdapt:25}, sparse-autoencoder concepts \citep{Tack:25}, or a hidden Markov model over continuous reasoning variables \citep{Liu:25}.
These methods share that intermediate computation is externalized by autoregressively generating continuous representations.

Other methods are at the intersection of explicit and implicit reasoning. For instance,
\citet{Kim:25} dynamically insert pause tokens at low-confidence positions, \citet{Sun:25} attach filler tokens to generated tokens, and \citet{Zhang:25} dynamically compress generated reasoning steps into continuous space.
While these methods reduce or redistribute the cost of explicit reasoning, they still operate within the standard autoregressive generation paradigm.

\paragraph{Implicit latent reasoning.}
Another line of work studies whether intermediate computation can be allocated to tokens without predefined semantic content, which we collectively refer to as filler tokens.
Prior work shows that making filler tokens useful for reasoning is difficult.
\citet{Lanham:23} find that simply adding filler tokens does not improve accuracy and can even reduce it in longer-context settings.
\citet{Pfau:24} then show that filler tokens can support computation on synthetic algorithmic tasks, but require specific dense supervision to converge.
\citet{Goyal:24} scale this to real-world downstream tasks, but find that gains arise mainly when filler tokens are used during both pretraining and finetuning.
More recently, DART shows that filler tokens can be trained without specific pretraining, but finetuning requires a two-pathway self-distillation framework with an auxiliary ``Reasoning Evolvement Module'' and multiple auxiliary losses \citep{Deng:24}.
Together, these results suggest that filler-token reasoning is possible, but requires a carefully designed training signal.
Our method improves upon this line of work by showing that filler tokens inside fixed memory blocks can be trained as an effective latent workspace for reasoning, which we describe next.

\section{Reasoning in Memory} \label{sec:method}

In this section, we introduce \emph{\fullmethod{} (\method{})}, our method that uses memory blocks as a latent workspace for implicit latent reasoning.
We first formalize the sequential generation paradigm underlying chain-of-thought (CoT) and explicit latent reasoning methods, and then present how we can train models to effectively decouple intermediate computation from autoregressive generation.

\paragraph{CoT reasoning.} Formally, let $\Bx=(x_1,\ldots,x_q)$ and $\By=(y_1,\ldots,y_a)$ denote the discrete token sequences of the input question and the final answer, respectively, and let $\Bw$
denote the parameters of an autoregressive language model. Standard
CoT reasoning scales test-time computation by generating a
sequence of intermediate reasoning steps
$\Br_{1:T}=(\Br_1,\ldots,\Br_T)$ before the final answer \citep{Wei:22}.
Each reasoning step $\Br_t$ is itself a discrete token sequence generated autoregressively.
If the full reasoning trace is flattened into a token sequence
$\Br=(r_1,\ldots,r_C)$, then each reasoning token is decoded from the language model's predictive distribution
$p_{\Bw}(r_i \mid \Bx, \Br_{<i})$.
Thus, generating the reasoning trace requires $C$ sequential decoding steps.

\paragraph{Explicit latent reasoning.} Latent reasoning methods such as Coconut \citep{Hao:25} replace written
reasoning steps with continuous representations. 
Instead of decoding intermediate reasoning steps $\Br_{1:T}$, the model
generates a sequence of continuous representations
$\Bz_{1:L}=(\Bz_1,\ldots,\Bz_L)$, with $\Bz_\ell\in\mathbb{R}^d$. 
Unlike discrete tokens, these continuous representations are not sampled from a
predictive distribution. 
Rather, each $\Bz_\ell$ is obtained from the hidden state at a chosen layer of the language model, conditioned on the question $\Bx$ and
the previously generated continuous representations $\Bz_{<\ell}$. The resulting representation is then fed back as the input embedding at the next decoding step.
Thus, explicit latent reasoning with continuous representations requires $L$ sequential decoding steps, typically with $L < C$.
However, since each representation $\Bz_\ell$ must be generated before $\Bz_{\ell+1}$ can condition on it, intermediate computation remains coupled to autoregressive generation.

\paragraph{Our method.}
\method{} builds on the central advantage of explicit latent reasoning, namely that intermediate computation need not be expressed in natural language, but can instead take place in a continuous space.
However, instead of explicitly generating continuous representations and feeding them back into the language model, \method{} moves this computation into fixed working memory.
Concretely, we provide the language model with $K$ fixed memory blocks $\Bm_{1:K}=(\Bm_1,\ldots,\Bm_K)$, where each memory block consists of a predefined sequence of special tokens
\begin{equation}
    \Bm_k =
    [\texttt{<b>}, \underbrace{\texttt{<m>},\ldots,\texttt{<m>}}_{M\ \mathrm{tokens}}, \texttt{</b>}] \ .
\end{equation}
The idea is that $\texttt{<m>}$ tokens form the latent workspace, while $\texttt{<b>}$ and $\texttt{</b>}$ delimit block boundaries. 
Although memory blocks could in principle be constructed from existing vocabulary tokens, we introduce dedicated special tokens to avoid overloading pretrained token semantics. 
Additionally, to minimize interference with the pretrained vocabulary, we freeze the embeddings of all existing tokens and update only the embeddings of the special tokens.
This isolates adaptation to the newly introduced special tokens, ensuring that improvements arise from learning to use the memory blocks rather than from altering pretrained token representations.
The special tokens are otherwise treated as standard input tokens and are processed by the language model in the usual way.
Unless stated otherwise, we fix $M=2$ in all experiments.
The fixed memory blocks $\Bm_{1:K}$ are appended to the input question $\Bx$ and serve as a latent workspace for intermediate computation. Since the memory blocks are fixed input tokens, the language model processes the entire augmented sequence $(\Bx, \Bm_{1:K})$ in a single forward pass. Thus, \method{} retains the central benefit of reasoning in continuous space while avoiding the sequential bottleneck of generating continuous representations one at a time.

\paragraph{Learning to reason in memory.} As memory blocks have no predefined computational role, eliciting useful latent computation is a central challenge. Without targeted supervision, the language model may simply ignore them or treat them as distracting context, rather than using them to hold and manipulate task-relevant information \citep{Lanham:23,Pfau:24,Goyal:24}.
Thus, effective reasoning in memory requires a carefully structured supervision signal that teaches the model how to use these blocks as working memory.

To combine training efficiency with dense supervision, \method{} uses a simple two-stage curriculum.
Both stages use the same simple objective, standard next-token prediction, but with different targets.
This avoids multiple pathways and auxiliary objectives used by distillation-based latent reasoning methods \citep{Jiang:25,Shen:25}, keeping optimization simple while still providing dense supervision over the latent workspace.
We now describe the objectives for the two stages.

\begin{figure}[b]
    \centering
    \includegraphics[width=\linewidth, trim = 0.0cm 0.0cm 0.0cm 0.0cm, clip]{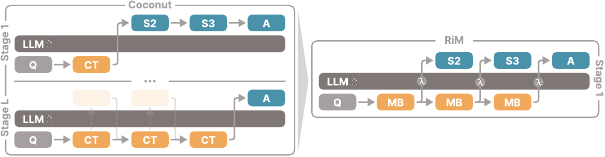}
    \caption{\textbf{\method{} Stage~1.}
    While Coconut \citep{Hao:25} uses multiple curriculum stages to train continuous thoughts (CTs), progressively increasing the number of steps, \method{} collapses this into a single stage over all memory blocks (MBs), forcing dense supervision through the latent workspace.}
    \label{fig:rim_stage_1}
\end{figure}

\subsection{Stage~1: Reasoning Step Supervision}

The goal of Stage~1 is to teach the model to use memory blocks as a latent workspace.
We achieve this by grounding them with explicit reasoning-step supervision. After each memory block, the model is trained to recover the next written reasoning step from the latent workspace available to it.
This encourages task-relevant information to be stored and transformed through the memory blocks.
Stage~1 is closely aligned with JEPA-style predictive representation learning because fixed memory blocks learn to predict missing reasoning structure, turning them into parallel latent states for working memory \citep{LeCun:22,Assran:23}.

A naive adaptation of Coconut-style staging \citep{Deng:24,Hao:25} would
gradually replace written reasoning steps with memory blocks.
In preliminary experiments, however, we found that this alone does not provide a sufficiently dense training signal.
First, during the early stages, target reasoning steps can attend to previous reasoning steps, allowing the model to predict the target directly from the preceding reasoning steps.
The supervision signal can therefore bypass the latent workspace, leaving the memory blocks with a weak training signal.
Second, Coconut-style staging requires fixing in advance a maximum number of training stages. It can therefore replace only that many reasoning steps with latent representations before switching to final-answer prediction. %
Consequently, some reasoning steps would never be directly supervised from the memory blocks.

\method{} avoids these shortcomings by replacing the written reasoning trace one-to-one
with memory blocks during Stage~1.
For a trace with $T$ reasoning steps, we allocate $T$ memory blocks and supervise
each block-level readout against the next reasoning step (\Cref{fig:rim_stage_1}).
To enable this dense supervision in a single forward pass, we employ a custom attention mask so each readout can attend to the memory blocks, and optionally to the question, but not to other
written reasoning steps (\Cref{fig:rim_attention_mask}). 
As a result, every reasoning step provides a direct supervision signal through the latent workspace. The model cannot recover it from previous written steps and must instead encode the necessary intermediate computation in the memory blocks themselves. We refer to \Cref{appendix:sec:training_details} for details.

Formally, for a training sample $(\Bx,\Br,\By)$ consisting of token sequences
corresponding to the question, reasoning trace, and final answer, respectively, we first segment the reasoning trace into $T$ reasoning steps
$\Br_{1:T}=(\Br_1,\ldots,\Br_T)$.
We then append one memory block per reasoning step and train the readout after
block $\Bm_t$ to predict the next reasoning step $\Br_{t+1}$, where we define
$\Br_{T+1}=\By$ for notational convenience.
The Stage~1 objective is a weighted reasoning-step negative log-likelihood,
\begin{equation}
    \mathcal{L}_{\mathrm{S1}}(\Bw)
    \, = \,
    - \sum_{t=1}^{T} \,
    \lambda_t(s) \,
    \log p_{\Bw}(\Br_{t+1} \mid \Bx,\Bm_{\le t}) \ ,
    \label{eq:stage1_loss}
\end{equation}
where $\lambda_t(s)\in[0,1]$ controls the supervision strength for the readout
after memory block $\Bm_t$, at training step $s$.
Annealing $\lambda_t(s)$ yields a soft multi-stage reasoning curriculum in which all readouts receive dense supervision early in training, before supervision is gradually removed from earlier reasoning steps first and later reasoning steps last.
We experimented with several weighting schedules and found that relative
weighting with respect to the number of reasoning steps $T$ in each sample
performs best.
Absolute weighting with respect to the maximum number of reasoning steps removes
supervision too early for shorter samples, while no weighting remains slightly worse
because dense supervision is never relaxed.
We therefore use a linear relative schedule throughout our experiments, where
$\lambda_t(s)$ decreases linearly from $1$ to $0$ over Stage~1, with the
annealing order determined by $t$.

\begin{figure}[b]
    \centering
    \includegraphics[width=\linewidth, trim = 0.14cm 0.0cm 0.04cm 0.0cm, clip]{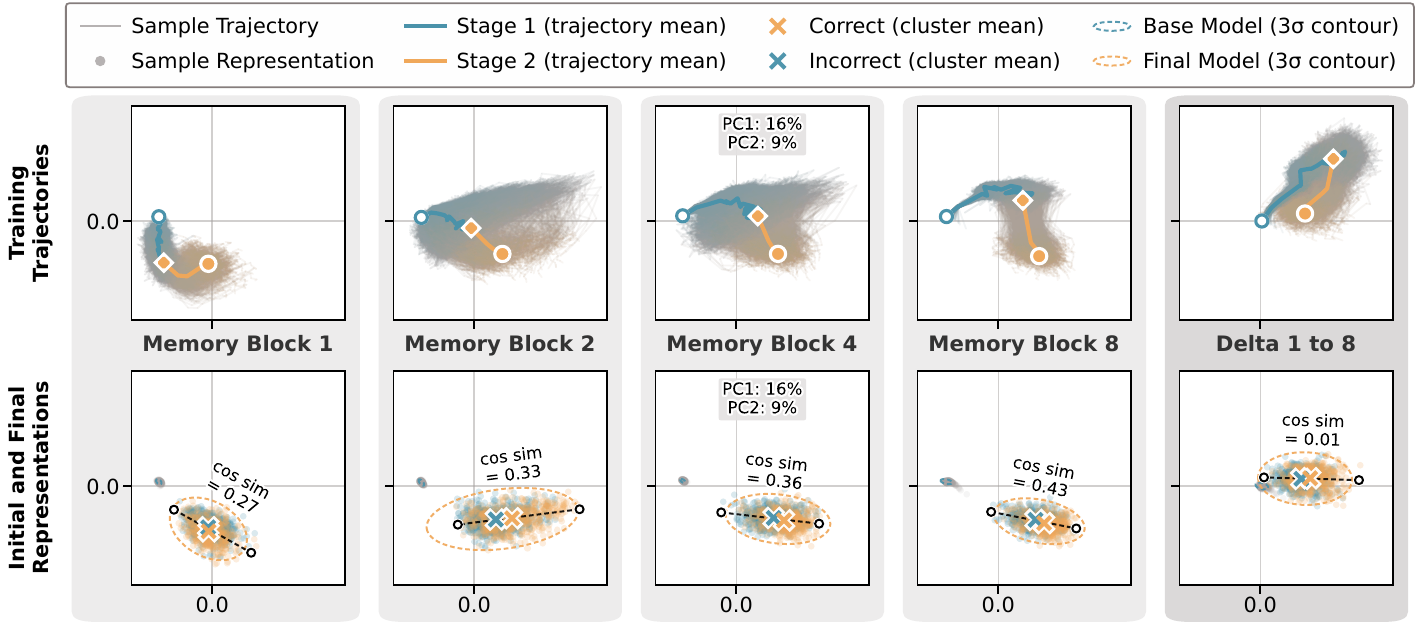}
    \caption{\textbf{Memory block representations.}
    Using all GSM8K test questions, we project memory block representations and the first-to-final memory block representation delta into a shared PCA basis.
    The top row shows their trajectories during training. The bottom row shows the representations from the initial base model (Llama-3.2-1B) and the final \method{}-trained model in the same PCA basis.}
    \label{fig:pca_trajectory_and_final_cloud_map}
\end{figure}

\subsection{Stage~2: Final Answer Refinement}

After Stage~1, the language model has been trained to route intermediate computation through memory blocks using written reasoning steps as supervision.
Stage~2 removes this intermediate supervision.
Instead of supervising each memory block to recover a reasoning step, the model is trained to directly predict and refine the final answer after each memory block.
We use the same custom causal mask as in Stage~1, so each readout can attend only to the question and the memory blocks available up to that point.
While Stage~1 uses one memory block per reasoning step to provide dense supervision, Stage~2 uses a fixed number of $K$ memory blocks and trains the model with an anytime answer objective, matching the inference-time setting more closely.
Each readout predicts the final answer from the memory blocks available up to that point, encouraging the model to use the learned latent workspace to improve its prediction as the available memory budget increases.

Formally, given the same training sample $(\Bx,\Br,\By)$, we discard the written reasoning trace $\Br$ and use a fixed number of $K$ memory blocks for every sample.
After each memory block $\Bm_k$, we attach an answer readout and train it to
predict the final answer $\By$.
The Stage~2 objective is a weighted final-answer negative log-likelihood,
\begin{equation}
    \mathcal{L}_{\mathrm{S2}}(\Bw)
    \, = \,
    - \sum_{k=1}^{K} \,
    \alpha_k \,
    \log p_{\Bw}(\By \mid \Bx,\Bm_{\le k}) \ ,
    \label{eq:stage2_loss}
\end{equation}
where $\alpha_k\in[0,1]$ controls the supervision strength for the readout after memory block $\Bm_k$, independent of the training step.
While this stage is relatively insensitive to the exact weighting schedule, we found that assigning larger weights to later memory blocks performs slightly better.
We again use a linear weighting schedule throughout our
experiments, reflecting that later readouts have access to more latent
computation and therefore should produce the strongest final answers.

Stage~2 is inspired by iterative latent reasoning models such as HRM \citep{WangHRM:25} and TRM
\citep{Martineau:25}, but realizes refinement horizontally along the
sequence dimension.
Additional memory blocks provide additional latent computation before the answer
readout, without requiring recurrent refinement modules or autoregressively
generated latent states.
In this sense, Stage~2 converts the grounded memory blocks from Stage~1 into a
fixed sequence of latent computations for final-answer prediction.

Following \citet{Deng:24,Hao:25}, we reset the optimizer state and
learning-rate scheduler at the stage switch.
We also use a lower learning rate and increased dropout to reduce
overfitting under dense answer supervision.
This hard switch is important because the two stages optimize distinct objectives:
Stage~1 grounds memory blocks for intermediate computation, whereas Stage~2 uses this learned latent workspace to refine the final answer under a fixed memory budget.

\section{Experiments} \label{sec:experiments}

Having introduced \fullmethod{} (\method{}), we now evaluate whether fixed memory blocks provide an effective mechanism for latent intermediate computation. We focus on three questions:
First, does \method{} induce the language model to use memory blocks for intermediate computation?
Second, how does \method{} compare to prior reasoning methods in terms of performance and latency?
Third, is the performance of \method{} robust across different inference-time memory budgets?

\paragraph{Models and Datasets.}
To answer these questions, we use the models and datasets adopted in established prior work on latent reasoning \citep{Deng:24,Hao:25,Shen:25,Goyal:24,Jiang:25}, enabling direct comparison with established methods.
For training, we use \emph{GSM8K-Aug} \citep{Deng:23}, a dataset of 386K grade-school math questions with up to 13 reasoning steps provided as mathematical expressions.
For evaluation, we use \emph{GSM8K} \citep{Cobbe:21} as the in-distribution (ID) test set and \emph{GSM-Hard} \citep{Gao:23} as an out-of-distribution (OOD) test set, consisting of more challenging math questions.
We evaluate across \emph{GPT-2} \citep{Radford:18} and \emph{Llama-3.2} \citep{Dubey:24} model scales, covering the two most common model families for latent reasoning.
Together, these evaluations cover transfer across model families, scaling across parameter sizes, and generalization from ID to OOD reasoning problems. 

\paragraph{Baselines.}
We compare \method{} against the most directly related reasoning baselines. 
First, we include supervised fine-tuning (SFT) \citep{Muennighoff:25} in two variants. \emph{SFT w/o CoT} is trained on question-answer pairs to directly predict the answer, making it the closest non-latent baseline to \method{}. \emph{SFT w/ CoT} is trained on explicit reasoning traces to generate a full CoT before predicting the answer, making it a strong explicit-reasoning baseline that incurs substantially higher inference cost.
Second, we include Coconut \citep{Hao:25}, the most widely used latent reasoning baseline, which replaces CoT reasoning with autoregressively generated CTs. \emph{Coconut w/ Stage~0} follows the original training recipe and gives Coconut a supervised warm start on explicit reasoning traces, while \emph{Coconut w/o Stage~0} omits this initial stage.
Third, we compare against DART \citep{Jiang:25} using the official reported numbers, since no official codebase is available and our reimplementation did not yield a reliable reproduction. This comparison favors DART in terms of training compute, since DART uses more epochs and a two-pathway training objective. Nevertheless, \method{} outperforms the reported DART results in all comparable settings, as discussed in \Cref{appendix:sec:training_details}.

\paragraph{Evaluation.} For SFT w/o CoT as well as all latent reasoning methods, we force the answer prefix ``\texttt{The final answer is \textbackslash boxed\{}'', ensuring they are evaluated only on final-answer generation.
In evaluating performance, prior work usually evaluates models at multiple training checkpoints and reports the checkpoint with the highest accuracy on the evaluation benchmark itself. This introduces selection overfitting and can overstate performance \citep{Cawley:10}. Thus, to separate selection from evaluation, we use $k$-fold cross-validation with a predefined checkpoint-selection protocol. For each of the 16 splits, we reserve 264 held-out GSM8K samples for checkpoint selection, choosing the checkpoint with the highest greedy accuracy for each model-method combination. Unless stated otherwise, these selected checkpoints are used throughout the following analyses to answer the three questions. %

\vspace{-2pt}
\subsection{\method{} Induces Latent Computation in Memory Blocks}
\label{subsec:q1}
\vspace{-2pt}
We first examine whether the language model learns nontrivial contextual representations at the memory-block positions.
If the model ignored the memory blocks, or used them only as fixed placeholders, their contextual representations should remain largely determined by token identity and position, with little systematic dependence on the input or on training.
By contrast, if the memory blocks served as a useful latent workspace, their contextual representations should change systematically during training and become input-dependent.
To assess which case applies, we train Llama-3.2-1B for 6 epochs in Stage~1 with up to 13 memory blocks and 2 epochs in Stage~2 with 8 memory blocks, saving checkpoints every 1,000 training steps.
This yields 18 and 6 checkpoints, respectively, at which we collect the penultimate-layer representation of each memory block per GSM8K test question.
We then project all representations into a shared PCA basis, which explains 25\% of the total variance.
\Cref{fig:pca_trajectory_and_final_cloud_map} shows the representations of individual memory blocks as well as the delta between the first and the final memory block.
For each question, this delta is computed in the original representation space before being projected into the same PCA basis, controlling for question-specific offsets in the representation space and thus isolating how the representations evolve across memory blocks rather than reflecting differences between questions.

\paragraph{Representation trajectories during training.}
In the top row of \Cref{fig:pca_trajectory_and_final_cloud_map}, we plot how the representations change during training, with a separate trajectory for each question as well as the mean trajectory of all questions.
We observe that these trajectories are block-specific and diverge as training progresses, which indicates that the model learns to make the latent workspace input dependent. Also, the mean trajectories are smooth, which indicates that training systematically organizes the representations rather than merely perturbing them.

\paragraph{Representations from base and final model.}
In the bottom row of \Cref{fig:pca_trajectory_and_final_cloud_map}, we plot the memory-block representations of the base model before training and the final model after training. 
These representations correspond to the start and end points of the trajectories shown in the top row.
While the base-model representations are largely collapsed, the final-model representations spread out and exhibit question-specific structure, suggesting that different questions induce distinct latent workspaces.
For the two samples that are farthest apart in the PCA projection, we additionally compute their cosine similarity in the original representation space. The low similarity confirms that the separation observed in the PCA projection reflects a corresponding difference in the original representations. This suggests that different questions induce distinct directions in the latent workspace, and that these differences are faithfully reflected by the PCA projection.
Together, these results show that memory blocks become both block-specific and sample-dependent, providing representation-level evidence that \method{} learns to use them as a latent workspace.
Plots for all 8 memory blocks and for representations from different layers of the language model are provided in \Cref{fig:appendix:pca_plots}.

\begin{figure}[t]
    \centering

    \begin{subfigure}[t]{0.49\linewidth}
        \centering
        \includegraphics[width=\linewidth, trim=0pt 0pt 0pt 0pt, clip]{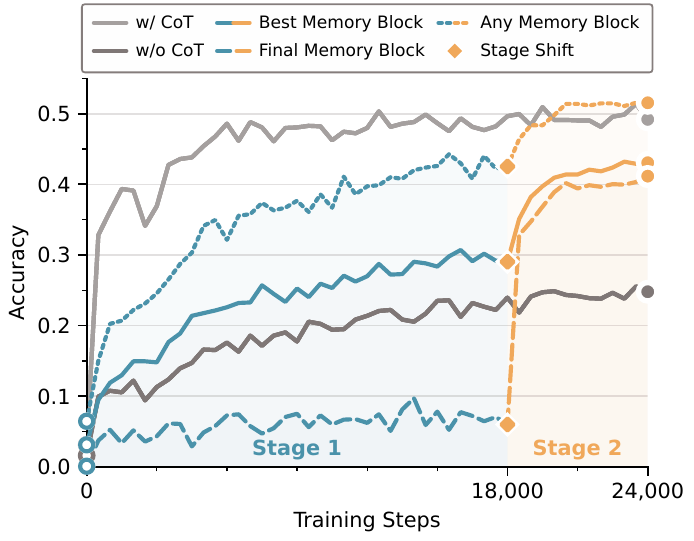}
        \caption{\method{} vs. SFT.}
        \label{fig:acc_sft}
    \end{subfigure}
    \hfill
    \begin{subfigure}[t]{0.49\linewidth}
        \centering
        \includegraphics[width=\linewidth, trim=0pt 0pt 0pt 0pt, clip]{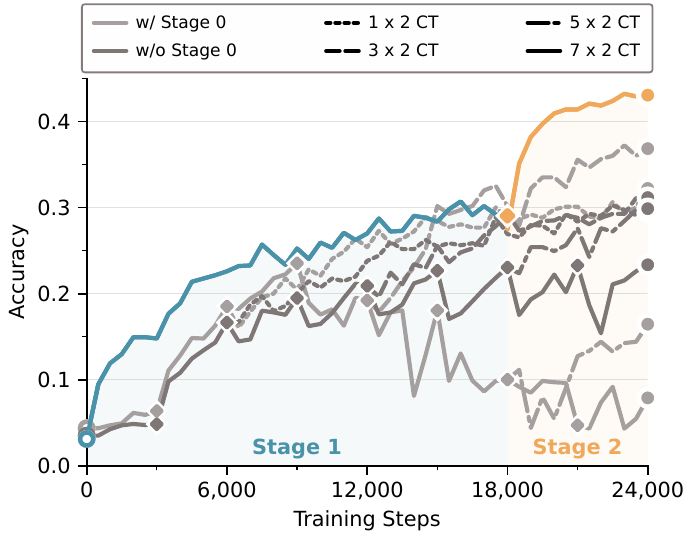}
        \caption{\method{} vs. Coconut.}
        \label{fig:acc_coconut}
    \end{subfigure}

    \caption{\textbf{Llama-3.2-1B training curves.}
    Greedy accuracy on GSM8K test questions over training, comparing \method{} to SFT and Coconut.
    \method{} is trained for 6 epochs in Stage~1 and 2 epochs in Stage~2, while Coconut is trained with 1 to 7 stages and 2 continuous thoughts (\texttt{x 2 CT}) added per stage.}
    
    \label{fig:acc_comp}
\end{figure}

\vspace{-2pt}
\subsection{\method{} Achieves High Performance at Low Latency}
\label{subsec:q2}
\vspace{-2pt}

Next, we examine whether the structured memory-block representations translate into strong final-answer accuracy and whether \method{} improves the accuracy-latency trade-off at inference time.

\paragraph{Performance during training.}
We first ask whether the structured memory-block representations observed above translate into improved task performance.
For \method{}, we distinguish three readouts. Final-block accuracy corresponds to the deployable fixed-budget setting used in the main comparison below, best-block accuracy reports the accuracy of the single best memory block, and any-block accuracy counts a sample as correct if any memory-block readout produces the correct answer.
\Cref{fig:acc_comp} tracks these readouts through training for Llama-3.2-1B on GSM8K.
\Cref{fig:acc_sft} compares \method{} to the two SFT variants.
The final-block readout improves sharply after the transition to Stage~2 and exceeds SFT w/o CoT, while any-block accuracy becomes competitive with SFT w/ CoT despite not generating explicit reasoning traces.
\Cref{fig:acc_coconut} compares \method{} to Coconut variants with different latent reasoning budgets.
Although additional latent stages give Coconut more sequential latent computation, they do not improve performance.
The strongest Coconut variant is the $3\times2$ configuration w/ Stage~0, which is also the configuration used in the original Coconut paper and in the main comparison below.
Here, $3\times2$ denotes 3 latent curriculum stages with 2 CTs added at each stage.
Under the same training budget, \method{} consistently outperforms all Coconut variants.

\paragraph{Main results.} 
\Cref{tab:main_results} shows that the deployable final-block readout of \method{} consistently outperforms the strongest Coconut variant, with improvements between $2.5$ and $7.5$ 
percentage points (pp) on GSM8K and between $0.7$ and $1.8$ 
pp on GSM-Hard. %
The gains over direct-answer SFT are even larger, with improvements between $12.6$ and $18.2$ pp on GSM8K and between $3.5$ and $5.2$ pp on GSM-Hard.
The pass@8 accuracies, obtained by sampling 8 answers with temperature 1, show the same pattern. \method{} samples correct solutions more reliably than the baselines across all benchmarks.
This shows that memory blocks do not merely introduce stochastic variation but create useful internal computation that assigns substantial probability mass to correct answers, even when the greedy output is not always correct.
Additionally, in \Cref{tab:main_results2}, we compare \method{} to SFT w/ CoT as a diagnostic reference for explicit reasoning. The results show that \method{} often reaches correct solutions along its latent trajectory, approaching or even exceeding explicit CoT at substantially lower inference latency. Since any-block accuracy assumes access to the best memory-block readout, we treat it as a measure of the performance potential of \method{} rather than as the main deployable result. Nonetheless, in \Cref{tab:appendix_probe_separability}, we show that linear probes on the memory block representations largely close the gap between final-block and any-block accuracy.

\begin{table*}[t]
\centering
\small
\setlength{\tabcolsep}{4.3pt}
\renewcommand{\arraystretch}{1.2}
\caption{\textbf{Main Results.} 
Accuracy ($\uparrow$) on GSM8K and GSM-Hard, with checkpoints selected using the cross-validation protocol on GSM8K.
Values report mean $\pm$ standard error over 16 split repeats.}
\begin{tabular}{lllrcccc}
\toprule
\multirow{2.3}{*}{Model} & \multirow{2.3}{*}{Method} & \multirow{2.3}{*}{Variant} & \multirow{3.23}{*}{\shortstack{TTFT\\(ms)}} &\multicolumn{2}{c}{GSM8K (ID)} & \multicolumn{2}{c}{GSM-Hard (OOD)} \\
\cmidrule(lr){5-6} \cmidrule(lr){7-8}
 & & & & Greedy (\%) & Pass@8 (\%) & Greedy (\%) & Pass@8 (\%) \\
\midrule
\multirow{3}{*}{GPT-2} & SFT & w/o CoT & \best{7.6} &
15.4{\scriptsize\,$\pm$\,0.2} & 33.3{\scriptsize\,$\pm$\,0.3} & 3.5{\scriptsize\,$\pm$\,0.1} & 7.6{\scriptsize\,$\pm$\,0.1} \\
 & Coconut & w/ Stage~0 & 53.4 &
 31.1{\scriptsize\,$\pm$\,0.2} & 45.0{\scriptsize\,$\pm$\,0.2} & 7.1{\scriptsize\,$\pm$\,0.0} & 10.7{\scriptsize\,$\pm$\,0.1} \\
 & \cellcolor{OursMainYellow}\method{} (ours) & \cellcolor{OursMainYellow}Final block & \cellcolor{OursMainYellow}\best{7.6} & 
 \cellcolor{OursMainYellow}\best{33.6}{\scriptsize\,$\pm$\,0.2} & \cellcolor{OursMainYellow}\best{49.1}{\scriptsize\,$\pm$\,0.2} & \cellcolor{OursMainYellow}\best{7.8}{\scriptsize\,$\pm$\,0.1} & \cellcolor{OursMainYellow}\best{11.2}{\scriptsize\,$\pm$\,0.1} \\
\midrule
\multirow{3}{*}{Llama-3.2-1B} & SFT & w/o CoT & \best{16.1} & 
23.9{\scriptsize\,$\pm$\,0.2} & 41.7{\scriptsize\,$\pm$\,0.3} & 5.3{\scriptsize\,$\pm$\,0.1} & 9.5{\scriptsize\,$\pm$\,0.1} \\
 & Coconut & w/ Stage~0 & 108.3 &
 36.9{\scriptsize\,$\pm$\,0.2} & 51.1{\scriptsize\,$\pm$\,0.2} & 8.5{\scriptsize\,$\pm$\,0.0} & 12.2{\scriptsize\,$\pm$\,0.0} \\
 & \cellcolor{OursMainYellow}\method{} (ours) & \cellcolor{OursMainYellow}Final block & \cellcolor{OursMainYellow}\best{16.1} &
 \cellcolor{OursMainYellow}\best{42.1}{\scriptsize\,$\pm$\,0.2} & \cellcolor{OursMainYellow}\best{56.1}{\scriptsize\,$\pm$\,0.3} & \cellcolor{OursMainYellow}\best{10.5}{\scriptsize\,$\pm$\,0.0} & \cellcolor{OursMainYellow}\best{13.8}{\scriptsize\,$\pm$\,0.0} \\
\midrule
 \multirow{3}{*}{Llama-3.2-3B} & SFT & w/o CoT & \best{27.9} & 
 36.2{\scriptsize\,$\pm$\,0.2} & 45.9{\scriptsize\,$\pm$\,0.2} & 8.5{\scriptsize\,$\pm$\,0.1} & 10.8{\scriptsize\,$\pm$\,0.1} \\
 & Coconut & w/ Stage~0 & 188.8 &
 41.3{\scriptsize\,$\pm$\,0.2} & 55.5{\scriptsize\,$\pm$\,0.5} & 10.2{\scriptsize\,$\pm$\,0.1} & 13.5{\scriptsize\,$\pm$\,0.1} \\
 & \cellcolor{OursMainYellow}\method{} (ours) & \cellcolor{OursMainYellow}Final block & \cellcolor{OursMainYellow}\best{27.9} &
 \cellcolor{OursMainYellow}\best{48.8}{\scriptsize\,$\pm$\,0.2} & \cellcolor{OursMainYellow}\best{58.8}{\scriptsize\,$\pm$\,0.2} & \cellcolor{OursMainYellow}\best{12.0}{\scriptsize\,$\pm$\,0.0} & \cellcolor{OursMainYellow}\best{14.1}{\scriptsize\,$\pm$\,0.0} \\
\bottomrule
\end{tabular}
\label{tab:main_results}
\end{table*}

\begin{table*}[b]
\centering
\small
\setlength{\tabcolsep}{4.3pt}
\renewcommand{\arraystretch}{1.2}
\caption{\textbf{Additional Results.}
Accuracy ($\uparrow$) on GSM8K and GSM-Hard for SFT w/ CoT and the any-block readout of \method{}.
Checkpoints are selected using the same cross-validation protocol on GSM8K as in the main results.
Values again report mean $\pm$ standard error over 16 split repeats.}
\begin{tabular}{lllrcccc}
\toprule
\multirow{2.3}{*}{Model} & \multirow{2.3}{*}{Method} & \multirow{2.3}{*}{Variant} & \multirow{3.23}{*}{\shortstack{TTFT\\(ms)}} & \multicolumn{2}{c}{GSM8K (ID)} & \multicolumn{2}{c}{GSM-Hard (OOD)} \\
\cmidrule(lr){5-6} \cmidrule(lr){7-8}
 &  &  & & Greedy (\%) & Pass@8 (\%) & Greedy (\%) & Pass@8 (\%) \\
\midrule
\multirow{2}{*}{GPT-2} & SFT & w/ CoT & 213.7 &
 \best{39.8}{\scriptsize\,$\pm$\,0.2} & 57.0{\scriptsize\,$\pm$\,0.2} & 8.4{\scriptsize\,$\pm$\,0.0} & 12.9{\scriptsize\,$\pm$\,0.1} \\
 & \cellcolor{OursMainYellow}\method{} (ours) & \cellcolor{OursMainYellow}Any block & \cellcolor{OursMainYellow}\best{7.6} &
 \cellcolor{OursMainYellow}39.5{\scriptsize\,$\pm$\,0.3} & \cellcolor{OursMainYellow}\best{78.1}{\scriptsize\,$\pm$\,0.2} & \cellcolor{OursMainYellow}\best{9.4}{\scriptsize\,$\pm$\,0.0} & \cellcolor{OursMainYellow}\best{19.0}{\scriptsize\,$\pm$\,0.1} \\
\midrule
\multirow{2}{*}{Llama-3.2-1B} & SFT & w/ CoT & 420.3 &
 49.1{\scriptsize\,$\pm$\,0.4} & 64.7{\scriptsize\,$\pm$\,0.3} & 11.2{\scriptsize\,$\pm$\,0.1} & 15.3{\scriptsize\,$\pm$\,0.1} \\
 & \cellcolor{OursMainYellow}\method{} (ours) & \cellcolor{OursMainYellow}Any block & \cellcolor{OursMainYellow}\best{16.1} &
 \cellcolor{OursMainYellow}\best{51.4}{\scriptsize\,$\pm$\,0.2} & \cellcolor{OursMainYellow}\best{76.8}{\scriptsize\,$\pm$\,0.1} & \cellcolor{OursMainYellow}\best{13.0}{\scriptsize\,$\pm$\,0.0} & \cellcolor{OursMainYellow}\best{19.6}{\scriptsize\,$\pm$\,0.1} \\
\midrule
\multirow{2}{*}{Llama-3.2-3B} & SFT & w/ CoT & 754.4 &
 \best{66.9}{\scriptsize\,$\pm$\,0.2} & \best{78.3}{\scriptsize\,$\pm$\,0.4} & \best{19.0}{\scriptsize\,$\pm$\,0.1} & \best{24.4}{\scriptsize\,$\pm$\,0.3} \\
 & \cellcolor{OursMainYellow}\method{} (ours) & \cellcolor{OursMainYellow}Any block & \cellcolor{OursMainYellow}\best{27.9} &
 \cellcolor{OursMainYellow}57.3{\scriptsize\,$\pm$\,0.1} & \cellcolor{OursMainYellow}71.8{\scriptsize\,$\pm$\,0.2} & \cellcolor{OursMainYellow}13.8{\scriptsize\,$\pm$\,0.1} & \cellcolor{OursMainYellow}18.2{\scriptsize\,$\pm$\,0.0} \\
\bottomrule
\end{tabular}
\label{tab:main_results2}
\end{table*}

\paragraph{Inference latency.} In \Cref{tab:main_results,tab:main_results2}, we also report the time to first token (TTFT) per sample in milliseconds (ms), averaged over all GSM8K test questions and four independent runs. It can be observed that \method{} has essentially the same TTFT as SFT w/o CoT. This is because the memory blocks are a fixed, small number of input tokens, processed in a single forward pass. 
By contrast, Coconut is about 7 times slower than \method{} due to its autoregressive nature. SFT w/ CoT is about 27 times slower than \method{}. These slowdowns are consistent across model scales.
Thus, \method{} has a favorable accuracy-latency tradeoff, improving substantially over direct-answer SFT and the strongest Coconut baseline while preserving direct-answer inference speed.

\vspace{-2pt}
\subsection{\method{} Maintains Accuracy Across Inference-Time Memory Budgets}
\label{subsec:q3}
\vspace{-2pt}
Finally, we examine whether \method{} maintains strong performance across inference-time memory budgets and how final answers evolve as more memory blocks are provided.

\begin{figure}[t]
    \centering

    \begin{subfigure}[t]{0.49\linewidth}
        \centering
        \includegraphics[width=\linewidth, trim=0pt 0pt 0pt 0pt, clip]{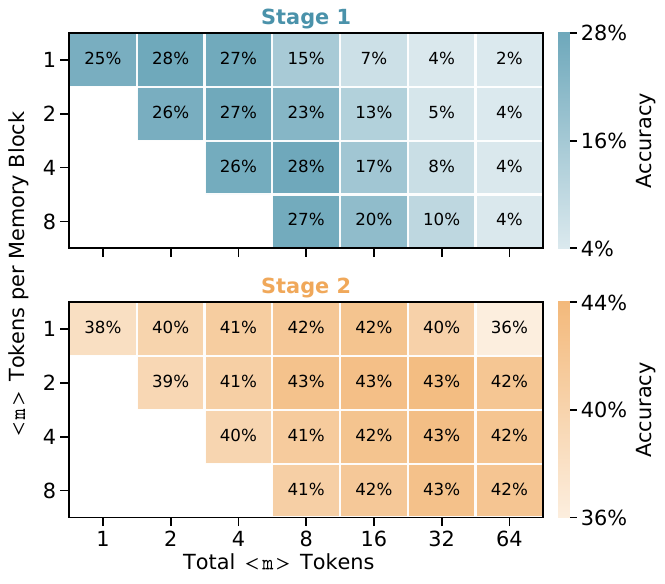}
        
        \caption{Accuracy across inference-time budgets.}
        \label{fig:stage_heatmaps}
    \end{subfigure}
    \hfill
    \begin{subfigure}[t]{0.49\linewidth}
        \centering
        \includegraphics[width=\linewidth, trim=0pt 0pt 0pt 0pt, clip]{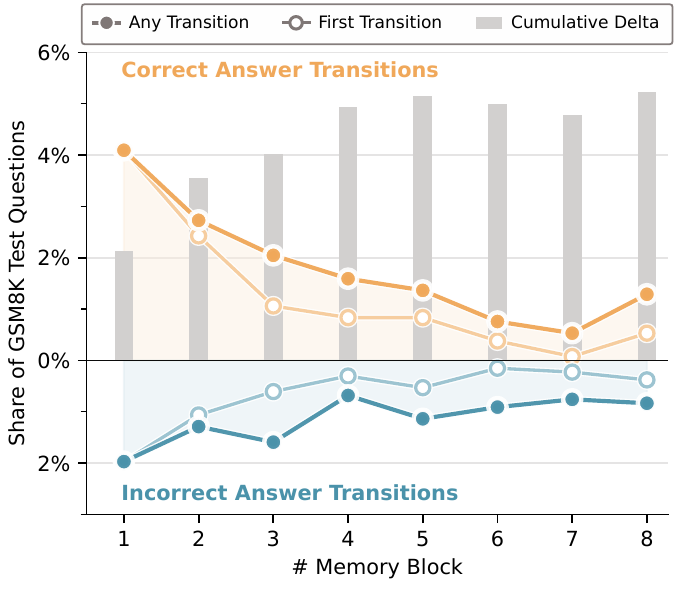}
        \caption{Answer transitions across memory blocks.}
        \label{fig:answer_transitions}
    \end{subfigure}

    \caption{\textbf{Robustness across memory budgets.}
    \method{}-trained Llama-3.2-1B on GSM8K test questions.
    \emph{(a)} Greedy accuracy for different numbers and sizes of memory blocks after both Stage~1 and Stage~2.
    \emph{(b)} Answer transition per memory block after Stage~2. Positive and negative values denote incorrect-to-correct and correct-to-incorrect changes, while gray bars show the cumulative net accuracy change.} %
    \label{fig:stage_performance}
    \vspace{-5pt}
\end{figure}

\paragraph{Performance across memory budgets.}
\Cref{fig:stage_heatmaps} evaluates Stage~1 and Stage~2 checkpoints while varying the number of \texttt{<m>} tokens per memory block ($M$) on the y-axis and the number of memory blocks ($K$) on the x-axis.
After Stage~1, greedy accuracy reaches around 27\% on GSM8K for the first memory-block readouts, but degrades when $K$ increases.
This is expected, since Stage~1 ties memory blocks to intermediate reasoning steps rather than directly to the final answer.
After Stage~2, however, this dependence largely disappears.
Accuracy increases to about 43\% and remains stable across a wide range of memory budgets.
This suggests that Stage~2 converts the grounded memory blocks learned in Stage~1 into a fixed sequence of latent computations that can be read out reliably across positions, making \method{}-trained models easy to deploy in practice. %

\paragraph{Answer transitions.}
We next track answer transitions across memory blocks to determine whether additional memory blocks continue to refine the final answer or whether the readouts collapse to the same prediction.
\Cref{fig:answer_transitions} shows that, across memory blocks, final answers of the final model still change for a nontrivial fraction of questions, with a positive cumulative net effect.
This suggests that the latent workspace does not collapse in Stage~2, but continues to refine predictions across memory blocks.
It shows that \method{} learns to use working memory in a stable yet nontrivial way, where additional memory blocks can change the answer to improve rather than destabilize performance.
Further experimental details are provided in \Cref{appendix:sec:training_details}, and additional results are reported in \Cref{appendix:sec:experimental_details}.

\vspace{-2pt}
\section{Conclusion}
\vspace{-2pt}
In this work, we introduce \fullmethod{} (\method{}), showing that large language models can reason in working memory rather than having to generate intermediate thoughts autoregressively.
Our method replaces reasoning steps with fixed memory blocks whose contextual representations form a latent workspace for task-specific computation.
Since these memory blocks are fixed, they decouple internal reasoning from autoregressive generation and can be processed in a single forward pass.

Our experiments show that the two-stage curriculum turns the latent workspace induced by the memory blocks into a structured, block-specific, and sample-dependent workspace.
This latent workspace improves performance over established baselines, remains robust across inference-time memory budgets, and preserves direct-answer inference speed.
Future work could study whether reinforcement learning with final-answer rewards in Stage~2 can further improve the latent workspace, and explore hybrid approaches that combine \method{} with explicit generation for more complex problems.

\bibliographystyle{plainnat} %
\bibliography{main}

\newpage
\appendix

\section{Acknowledgments}
\vspace{-2pt}
The ELLIS Unit Linz, the LIT AI Lab, the Institute for Machine Learning, are supported by the Federal State Upper Austria. We thank the projects FWF AIRI FG 9-N (10.55776/FG9), AI4GreenHeatingGrids (FFG- 899943), Stars4Waters (HORIZON-CL6-2021-CLIMATE-01-01), FWF Bilateral Artificial Intelligence (10.55776/COE12). We thank NXAI GmbH, Audi AG, Silicon Austria Labs (SAL), Merck Healthcare KGaA, GLS (Univ. Waterloo), T\"{U}V Holding GmbH, Software Competence Center Hagenberg GmbH, dSPACE GmbH, TRUMPF SE + Co. KG. Lukas would also like to thank Stefan Arnreiter for his encouragement during this work in the spirit of NIPSILD.

\section{Broader Impact}
\label{appendix:sec:broader_impacts}
\vspace{-2pt}
This work studies latent reasoning methods for language models on mathematical reasoning benchmarks. A potential positive impact is improved inference efficiency for reasoning systems, since \method{} aims to replace autoregressively generated intermediate thoughts with fixed memory blocks that can be processed in a single forward pass. More efficient reasoning could reduce computational cost and energy use for applications that require multi-step problem solving. At the same time, moving intermediate computation into latent memory blocks can make the reasoning process less directly interpretable than written chain-of-thought traces. Our experiments do not release a new pretrained model or dataset, and they are limited to existing mathematical reasoning benchmarks.

\vspace{-2pt}
\section{Assets and Licenses}
\label{appendix:sec:asset_licenses}
\vspace{-2pt}

We use existing public datasets, pretrained model families, and baseline implementations, all of which are cited in the main text.
The GSM8K dataset \citep{Cobbe:21} is listed under the MIT License.\footnote{\url{https://huggingface.co/datasets/openai/gsm8k}}
GSM-Hard \citep{Gao:23} is listed under the MIT License.\footnote{\url{https://huggingface.co/datasets/reasoning-machines/gsm-hard}}
For GSM8K-Aug \citep{Deng:23}, our experiments use the \texttt{zen-E/GSM8k-Aug} Hugging Face distribution, which is listed under Apache-2.0.\footnote{\url{https://huggingface.co/datasets/zen-E/GSM8k-Aug}}
Regarding pretrained language models, Llama-3.2-1B and Llama-3.2-3B \citep{Dubey:24} are governed by the Llama 3.2 Community License and Acceptable Use Policy,\footnote{\url{https://huggingface.co/meta-llama/Llama-3.2-1B}}$^,$\footnote{\url{https://huggingface.co/meta-llama/Llama-3.2-3B}} while GPT-2 \citep{Radford:18} is released by OpenAI under a Modified MIT License\footnote{\url{https://github.com/openai/gpt-2}}.
Our Coconut baseline implementation is based on the official Coconut codebase \citep{Hao:25}, which is released under the MIT License.\footnote{\url{https://github.com/facebookresearch/coconut}}

\vspace{-2pt}
\section{Experimental Details} 
\label{appendix:sec:training_details}
\vspace{-2pt}

\begin{table*}[t]
\centering
\small
\setlength{\tabcolsep}{10pt}
\renewcommand{\arraystretch}{0.96}
\caption{\textbf{GSM8K-Aug composition.} Distribution of training samples by number of reasoning steps.}
\begin{tabular}{c r r}
\toprule
\textbf{Reasoning Steps} & \textbf{Training Samples} & \textbf{Share} \\
\midrule
1  & 62,908  & 16 \% \\
2  & 143,578 & 37 \% \\
3  & 104,249 & 27 \% \\
4  & 48,198  & 13 \% \\
5  & 17,906  & 5 \% \\
6  & 5,666   & 1 \% \\
7  & 2,359   & 1 \% \\
8  & 577     & 0 \% \\
9  & 126     & 0 \% \\
10 & 43      & 0 \% \\
11 & 8       & 0 \% \\
12 & 1       & 0 \% \\
13 & 1       & 0 \% \\
\midrule
\textbf{Total}  & \textbf{385,620} & \\
\bottomrule
\end{tabular}
\label{tab:appendix:training_steps_distribution}
\vspace{-3pt}
\end{table*}

All experiments were run on one node with 8 NVIDIA H200-SXM-144GB GPUs. %

\paragraph{Dataset Details.} \Cref{tab:appendix:training_steps_distribution} summarizes the reasoning-step distribution of GSM8K-Aug, which determines the maximum Stage~1 memory-block depth and the number of update steps per epoch.
The distribution is concentrated on short traces, but includes samples with up to 13 reasoning steps.
\Cref{fig:appendix:dataset_examples} shows representative samples from the training and out-of-distribution evaluation datasets.
For GSM8K-Aug, we use the explicit arithmetic reasoning steps for Stage~1 supervision, whereas for GSM-Hard, we only use the question and final target answer at evaluation time.

\paragraph{Baselines.} We compare against supervised fine-tuning (SFT) baselines, Coconut \citep{Hao:25} as the explicit latent reasoning method, and the official DART numbers \citep{Jiang:25}:
\begin{itemize}[leftmargin=*, topsep=2pt, itemsep=2pt, partopsep=0pt, parsep=0pt]

    \item \textbf{SFT w/o CoT.} The model is trained and evaluated directly from question to answer.
    This is the closest non-latent control for our forced-answer setting, since, like \method{}, it cannot write explicit reasoning traces at test time.
    
    \item \textbf{SFT w/ CoT.} The model is trained and evaluated with full natural-language reasoning traces. This is not computationally equivalent to \method{}, because it externalizes intermediate reasoning in text, but it provides the gold-standard reference for performance when explicit reasoning is allowed.
    
    \item \textbf{Coconut w/o Stage~0.} 
    \citet{Hao:25} replace the explicit CoT with continuous thoughts (CTs), feeding previous hidden states back as the next input embedding instead of decoding them into word tokens.
    The Coconut curriculum progressively removes early written reasoning steps and inserts CTs, while training the model to predict the remaining reasoning trace and final answer. This variant omits Stage~0, the initial training phase on explicit reasoning traces equivalent to SFT w/ CoT. It therefore starts directly from the latent curriculum, making it the most direct latent reasoning comparison to \method{} at the Llama model scales.

    \item \textbf{Coconut w/ Stage~0.} This variant follows the original Coconut recipe more closely by adding Stage~0 and therefore using a larger effective training budget. We compare against this stronger variant in our main results in \Cref{tab:main_results} across all model scales. Our Coconut implementation is based on the official codebase \citep{Hao:25}.

    \item \textbf{DART.} We attempted to reimplement DART, but to the best of our knowledge, no official codebase has been made public to date, and we could not obtain a reliable reproduction. We therefore compare to DART using the official results reported by \citet{Jiang:25} in \Cref{appendix:tab:official_comparison}.
    This comparison is conservative for \method{} in terms of training cost, since the official DART results train Llama-3.2-1B and Llama-3.2-3B for 10 epochs each, and GPT-2 for 40 epochs.
    Since DART also requires two training pathways, this corresponds to a significantly higher training cost than \method{}.
    Nevertheless, \method{} outperforms the reported DART results in all comparable settings (see \Cref{appendix:tab:official_comparison}).

\end{itemize}

\begin{figure*}[!t]
\centering

\begin{subfigure}[t]{\linewidth}
\centering
\begin{tcolorbox}[
    width=\linewidth,
    colback=gray!5,
    colframe=gray!50,
    boxrule=0.5pt,
    arc=2mm,
    left=2.4mm,
    right=0mm,
    top=2.4mm,
    bottom=2mm,
]
\small
\textbf{Question:}
\begin{quote}
John begins his hike at 8:30 AM and finishes at 6:30 PM. He rests for 20 minutes at noon, 15 minutes in the afternoon and 30 minutes before finishing his hike. How many hours did he spend hiking?
\end{quote}

\textbf{Reasoning Steps:}
\[
\begin{aligned}
\texttt{<<} 18.5 - 8.5   &\texttt{ = }10 \texttt{>>} \\
\texttt{<<} 20 + 15 + 30 &\texttt{ = }65 \texttt{>>} \\
\texttt{<<} 65 / 60      &\texttt{ = }1.08333 \texttt{>>} \\
\texttt{<<} 10 - 1.08333 &\texttt{ = }8.91667 \texttt{>>}
\end{aligned}
\]

\textbf{Answer:}
\begin{quote}
The final answer is \boxed{8.92}
\end{quote}
\end{tcolorbox}
\caption{GSM8K-Aug training sample with a question, four reasoning steps, and the final target answer.}
\end{subfigure}

\vspace{0.75em}

\begin{subfigure}[t]{\linewidth}
\centering
\begin{tcolorbox}[
    width=\linewidth,
    colback=gray!5,
    colframe=gray!50,
    boxrule=0.5pt,
    arc=2mm,
    left=2.4mm,
    right=0mm,
    top=2.4mm,
    bottom=2mm,
]
\small
\textbf{Question:}
\begin{quote}
Mrs. Cruz is looking for a house that will not go beyond her \$400 000 budget. She saw a property that has a selling price of \$350 000. On top of that, the buyer has to pay a brokerage fee which is 5\% of the selling price, and also the transfer fee that is 12\% of the selling price. How much more is the total price of the house than Mrs. Cruz's budget?
\end{quote}

\textbf{Answer:}
\begin{quote}
The final answer is \boxed{9500.0}
\end{quote}
\end{tcolorbox}
\caption{GSM-Hard test sample with a question and the final target answer.}
\end{subfigure}

\caption{\textbf{Dataset samples.}
Random samples from the training and test datasets that were used throughout our experiments.}
\label{fig:appendix:dataset_examples}
\end{figure*}

\paragraph{\method{}.}
We use 2 $\texttt{<m>}$ tokens per memory block, matching Coconut, which uses 2 additional CTs at each stage. Since GSM8K-Aug contains samples with up to 13 reasoning steps (see \Cref{tab:appendix:training_steps_distribution}), Stage~1 uses up to 13 memory blocks, one for each reasoning step. In Stage~2, the reasoning-step targets are discarded, and only the final answer is supervised, so the number of memory blocks no longer has to match the number of reasoning steps. We therefore fix the number of memory blocks to 8, yielding 32 special tokens in total when including the block-boundary tokens. We choose this budget to align with \citet{Jiang:25}, who find that 30 silent tokens are optimal for their method, enabling a fair comparison.
We train for 6 epochs in Stage~1, corresponding to 18{,}000 training steps, and 2 epochs in Stage~2, corresponding to 6{,}000 training steps. For GPT-2, we adopt the same initial Stage~0 used for Coconut w/ Stage~0. For the Llama models, we start directly from the base checkpoints without Stage~0.

\paragraph{Hyperparameters.}
We train all models with rank-128 LoRA adapters using bfloat16 precision \citep{Hu:22}. For all methods, we use a global batch size of 128, resulting in about 3,000 update steps per epoch on GSM8K-Aug \citep{Deng:23}. This training setup largely follows prior work on latent reasoning \citep{Hao:25, Jiang:25, Shen:25}. We use a constant learning-rate schedule with a 4\% warmup period and a weight decay of 0.01 by default. In initial experiments, we found that training behavior was largely insensitive to the number of warmup steps and to moderate changes in weight decay. We therefore keep these parameters fixed across all experiments.
In contrast, we found the learning rate to have a substantial impact on training behavior. 
Thus, we perform a separate learning-rate search for each method-model combination over the range $[10^{-5}, 10^{-3}]$ for a fair comparison.
For Coconut, we additionally search over the maximum number of stages on Llama-3.2-1B and use the best-performing variant for all main comparisons (see \Cref{fig:acc_coconut}).

\begin{figure}[b]
    \centering
    \vspace{-5pt}
    \includegraphics[width=0.42\linewidth, trim = 0.0cm 0.0cm 0.0cm 0.0cm, clip]{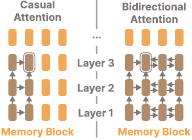}
    \caption{\textbf{Bidirectional memory block attention.}
    In this variant, tokens inside the same memory block can attend to each other bidirectionally, while attention between memory blocks remains causal.}
    \label{fig:appendix:bidirectional_attention_mask}
\end{figure}

\paragraph{Custom Attention Mask.}
We train with a custom block-causal attention mask that separates the sequence into a memory block stream and supervised written reasoning step readout branches.
Future memory blocks may attend to the question and previous memory blocks, but never to supervised reasoning step readouts.
Readouts may attend to the question and the memory blocks available up to their position, but not to other readouts.
This allows all reasoning steps or answer targets to receive teacher-forced supervision in a single forward pass while preventing later latent states from reading the ground-truth reasoning step.
In the default setting used throughout our experiments, tokens within each memory block remain causally ordered.
However, one can also consider the variant shown in \Cref{fig:appendix:bidirectional_attention_mask}, where tokens within a memory block attend bidirectionally to each other.
This increases within-block communication while preserving the same block-level causal structure.
Empirically, however, the bidirectional variant yields mixed results, with no consistent trend across models or benchmarks.
We therefore leave a systematic study of within-block attention structure to future work.

\pagebreak

\vspace{-2pt}
\section{Additional Results} 
\label{appendix:sec:experimental_details}
\vspace{-2pt}

\begin{figure}[t]
    \centering

    \begin{subfigure}[t]{0.46\linewidth}
        \centering
        \includegraphics[width=\linewidth, trim=5pt 0pt 5pt 0pt, clip]{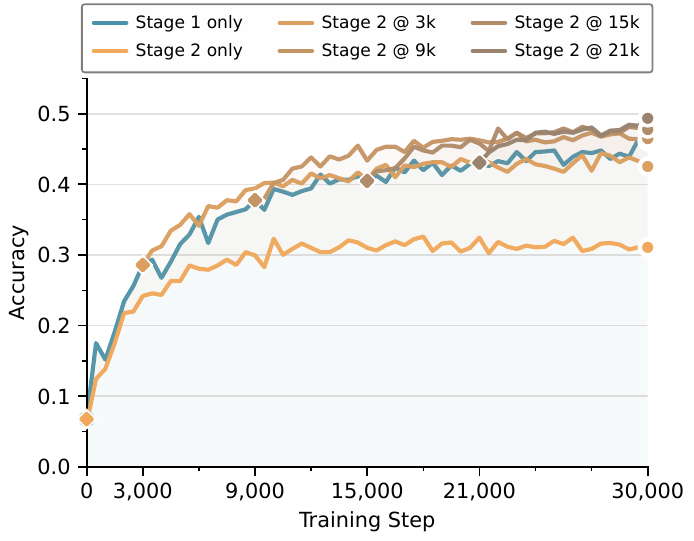}
        \caption{Any-block accuracy.}
        \label{fig:appendix:any_acc}
    \end{subfigure}
    \hfill
    \begin{subfigure}[t]{0.46\linewidth}
        \centering
        \includegraphics[width=\linewidth, trim=5pt 0pt 5pt 0pt, clip]{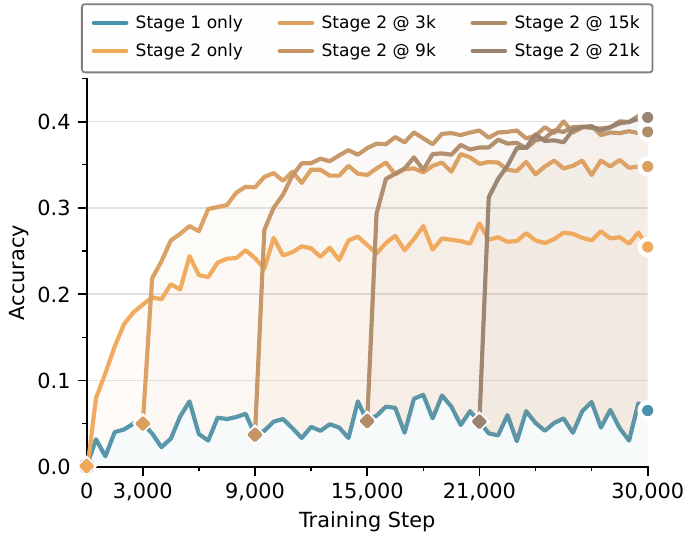}
        \caption{Final-block accuracy.}
        \label{fig:appendix:final_acc}
    \end{subfigure}

    \vspace{6pt}

    \begin{subfigure}[t]{0.46\linewidth}
        \centering
        \includegraphics[width=\linewidth, trim=5pt 0pt 5pt 0pt, clip]{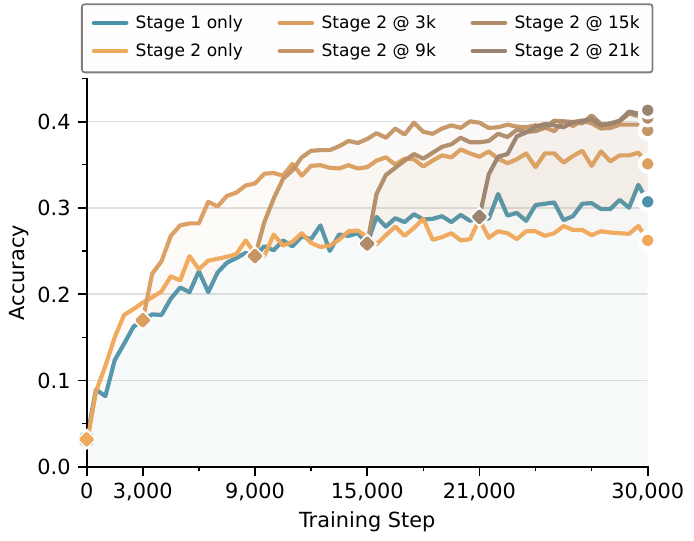}
        \caption{Best-block accuracy.}
        \label{fig:appendix:best_acc}
    \end{subfigure}

    \caption{\textbf{Stage-switch ablation.} GSM8K evaluation accuracy for Llama-3.2-1B trained on GSM8K-Aug, varying the stage switch from Stage~1 to Stage~2.}
    \label{fig:appendix:acc_comp}
    \vspace{-10pt}
\end{figure}

\paragraph{Stage-switch ablation.}
\Cref{fig:appendix:acc_comp} ablates the transition from Stage~1 to Stage~2.
Training with Stage~2 alone improves final-answer accuracy quickly, but plateaus well below runs that first ground the memory blocks with Stage~1. This supports the claim that dense subversion signal is important.
Conversely, Stage~1 alone produces high any-block accuracy, but its final-block accuracy remains low because the model has not been trained to use a fixed final readout. This supports the claim that switching to Stage~2 after sufficient Stage~1 training is important. As a result, the latent workspace learned in Stage~1 is converted into stronger final-block and best-block accuracy.
Very early switches are weaker, indicating that the memory blocks must first acquire a useful computational role before answer-only refinement becomes effective.

\begin{wrapfigure}{r}{0.46\textwidth}
    \vspace{-21pt}
    \centering
    \includegraphics[width=\linewidth, trim=0.0cm 0.0cm 0.0cm 0.0cm, clip]{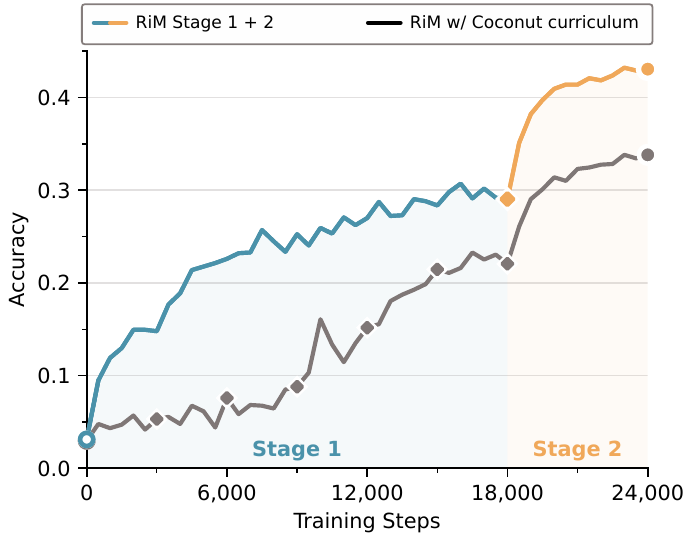}
    \caption{\textbf{\method{} vs. Coconut curriculum.} }
    \vspace{-20pt}
    \label{fig:coconut_curriculum_ablation}
\end{wrapfigure}

\paragraph{\method{} vs. Coconut curriculum.} 
To isolate the effect of our two-stage curriculum as presented in \Cref{sec:method}, we compare it with the staged curriculum used for Coconut \citet{Hao:25}, which was inspired by \citet{Deng:24}. This ablation keeps the fixed memory blocks of \method{}, but replaces our dense supervision signal with the gradual Coconut-style curriculum. \Cref{fig:coconut_curriculum_ablation} shows that the Coconut curriculum remains substantially below the \method{} curriculum across training steps. This confirms that memory blocks require a dense supervision signal that forces intermediate computation through the latent workspace in order to achieve high downstream performance.

\pagebreak
\begin{figure}[t]
    \centering

    \begin{subfigure}[t]{\linewidth}
        \centering
        \includegraphics[width=\linewidth, trim=0.14cm 0pt 0.04cm 0pt, clip]{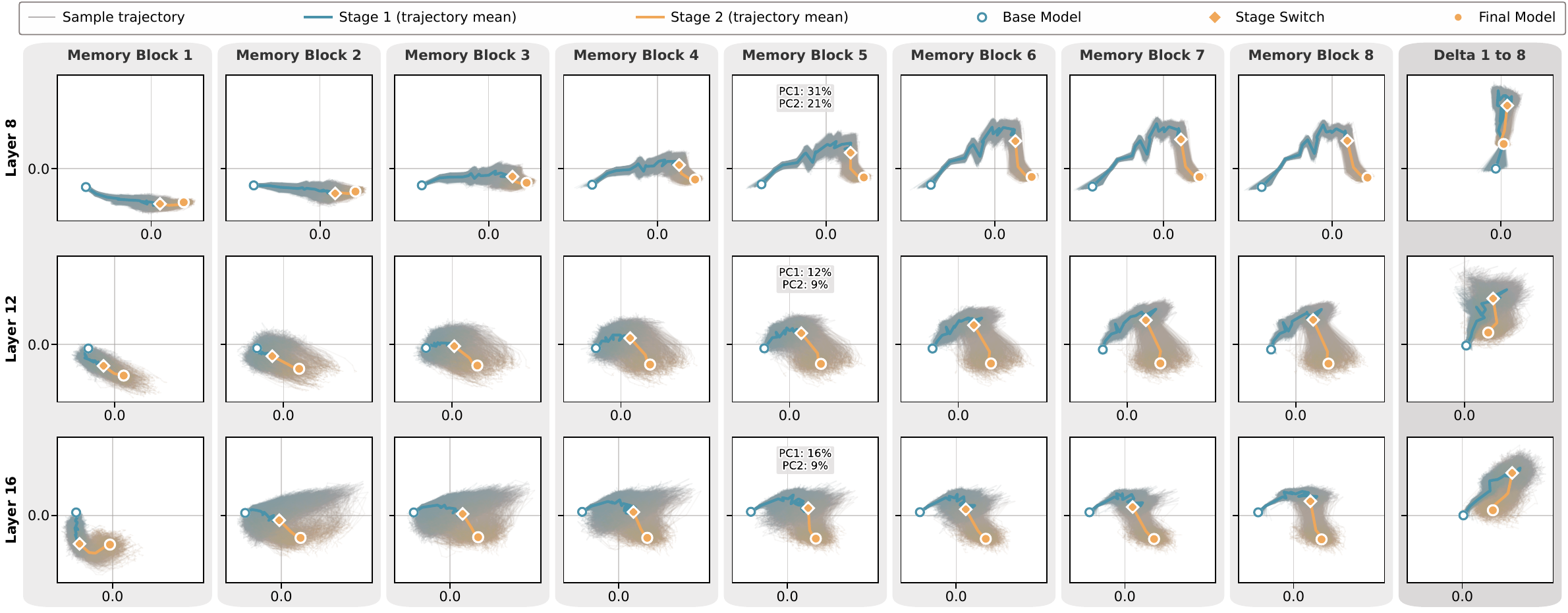}
        \caption{Training Trajectories.}
        \label{fig:appendix:training_traj}
    \end{subfigure}

    \vspace{12pt}
    
    \hfill
    \begin{subfigure}[t]{\linewidth}
        \centering
        \includegraphics[width=\linewidth, trim=0.14cm 0pt 0.04cm 0pt, clip]{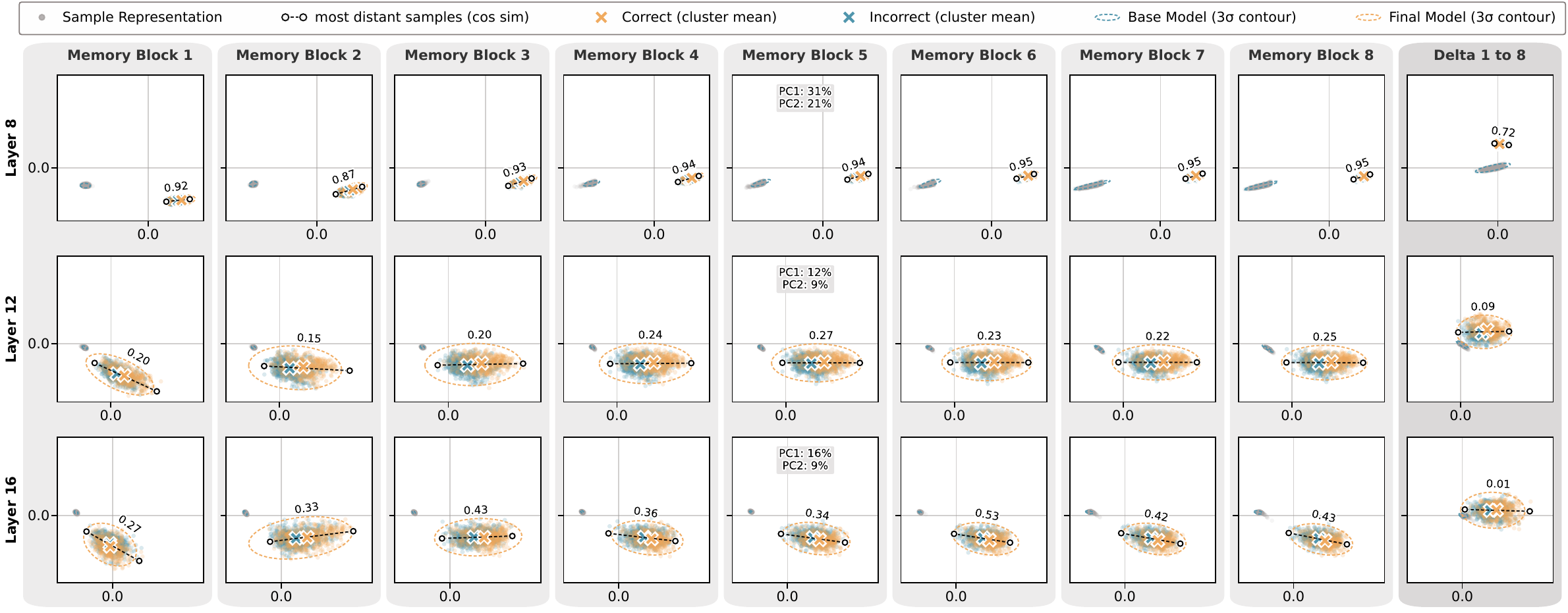}
        \caption{Initial and Final Representations.}
        \label{fig:appendix:final_rep}
    \end{subfigure}

    \caption{\textbf{Memory block representations.} Using all GSM8K test questions, we project memory block representations and the first-to-final memory block representation delta into a shared PCA basis. The top row shows their trajectories during training. The bottom row shows the representations from the initial base model (Llama-3.2-1B) and the final \method{}-trained model in the same PCA basis.}
    \label{fig:appendix:pca_plots}
\end{figure}

\paragraph{Latent computation in memory blocks.}
\Cref{fig:appendix:pca_plots} expands the representation analysis from \Cref{fig:pca_trajectory_and_final_cloud_map} of the main paper. For each layer, we collect the contextual representation of the last \texttt{<m>} token in each memory block across all GSM8K test questions and project them from both stages into a shared PCA basis. The same basis is used for the first-to-final memory block representation delta, so panels within each row are directly comparable.

The training trajectories in \Cref{fig:appendix:training_traj} show that memory block representations move along smooth and block-specific trajectories. Stage~1 already organizes the latent workspace, while Stage~2 changes the direction of the trajectories rather than collapsing them. 
This matches the objective shift, as Stage~1 grounds the memory blocks with reasoning-step supervision, whereas Stage~2 removes this strict step-level supervision and converts the grounded memory blocks into a fixed sequence of latent computations for final-answer prediction.

\Cref{fig:appendix:final_rep} compares the initial base model with the final \method{} model in the same PCA bases. The base-model representations are largely collapsed, while the final representations form broad, sample-dependent clouds whose locations remain organized by memory block. 
The first-to-final delta makes this especially clear, as in later layers, the delta representations become highly diverse, with the annotated most-distant-sample cosine similarity dropping to $0.09$ in Layer~12 and $0.01$ in Layer~16.
Thus, the memory blocks are not used as fixed placeholders but become structured, block-specific, and sample-dependent latent states. The figures provide representation-level evidence that \method{} trains the model to use memory blocks as a latent workspace for task-relevant intermediate computation.

\pagebreak

\paragraph{Detailed main results.}
\Cref{tab:appendix_main_results} expands the main results from \Cref{tab:main_results} by reporting all baseline variants and an additional ablation for \method{}.
The final-block row corresponds to the deployable setting used in the main paper and remains the strongest answer-only method across all backbones and benchmarks.
The any-block row asks whether any memory-block readout produces the correct answer, and therefore cannot be interpreted as a deployable selection rule.
Its large gains, especially for pass@8, show that the sequence of memory-block readouts often contains correct solutions even when the final-block readout is incorrect under deterministic decoding.
This suggests that \method{} does more than improve a single readout, but instead creates a sequence of latent states from which useful candidate solutions become available at different depths.

\begin{table*}[t]
\centering
\small
\setlength{\tabcolsep}{4.3pt}
\renewcommand{\arraystretch}{1.2}
\caption{\textbf{Main Results.} Accuracy ($\uparrow$) on two evaluation benchmarks, with checkpoints selected using the cross-validation protocol on GSM8K. Values report mean $\pm$ standard error over 16 split repeats. Bold indicates the best result, and underline indicates the second-best result under each model.}
\begin{tabular}{lllrcccc}
\toprule
\multirow{2.3}{*}{Model} & \multirow{2.3}{*}{Method} & \multirow{2.3}{*}{Variant} & \multirow{3.23}{*}{\shortstack{TTFT\\(ms)}} & \multicolumn{2}{c}{GSM8K (ID)} & \multicolumn{2}{c}{GSM-Hard (OOD)} \\
\cmidrule(lr){5-6} \cmidrule(lr){7-8}
 &  &  & & Greedy (\%) & Pass@8 (\%) & Greedy (\%) & Pass@8 (\%) \\
\midrule
\multirow{6}{*}{GPT-2} & \multirow{2}{*}{SFT} & w/o CoT & \best{7.6} &
15.4{\scriptsize\,$\pm$\,0.2} & 33.3{\scriptsize\,$\pm$\,0.3} & 3.5{\scriptsize\,$\pm$\,0.1} & 7.6{\scriptsize\,$\pm$\,0.1} \\
 &  & w/ CoT & 213.7 &
 \best{39.8}{\scriptsize\,$\pm$\,0.2} & \underline{57.0}{\scriptsize\,$\pm$\,0.2} & \underline{8.4}{\scriptsize\,$\pm$\,0.0} & \underline{12.9}{\scriptsize\,$\pm$\,0.1} \\
\noalign{\vskip 2pt}
\cdashline{2-8}
\noalign{\vskip 2pt}
 & \multirow{2}{*}{Coconut} & w/o Stage~0 & \multirow{2}{*}{53.4} &
 22.3{\scriptsize\,$\pm$\,0.3} & 40.6{\scriptsize\,$\pm$\,0.3} & 5.2{\scriptsize\,$\pm$\,0.0} & 8.9{\scriptsize\,$\pm$\,0.2} \\
 &  & w/ Stage~0 &  &
 31.1{\scriptsize\,$\pm$\,0.2} & 45.0{\scriptsize\,$\pm$\,0.2} & 7.1{\scriptsize\,$\pm$\,0.0} & 10.7{\scriptsize\,$\pm$\,0.1} \\
\noalign{\vskip 2pt}
\cdashline{2-8}
\noalign{\vskip 2pt}
 & \cellcolor{OursMainYellow} & \cellcolor{OursMainYellow}Final block & \cellcolor{OursMainYellow} & \cellcolor{OursMainYellow}33.6{\scriptsize\,$\pm$\,0.2} & \cellcolor{OursMainYellow}49.1{\scriptsize\,$\pm$\,0.2} & \cellcolor{OursMainYellow}7.8{\scriptsize\,$\pm$\,0.1} & \cellcolor{OursMainYellow}11.2{\scriptsize\,$\pm$\,0.1} \\
 & \cellcolor{OursSoftYellow}\multirow{-2}{*}{\method{} (ours)} & \cellcolor{OursSoftYellow}Any block & \cellcolor{OursSoftYellow}\multirow{-2}{*}{\best{7.6}} &
 \cellcolor{OursSoftYellow}\underline{39.5}{\scriptsize\,$\pm$\,0.3} & \cellcolor{OursSoftYellow}\best{78.1}{\scriptsize\,$\pm$\,0.2} & \cellcolor{OursSoftYellow}\best{9.4}{\scriptsize\,$\pm$\,0.0} & \cellcolor{OursSoftYellow}\best{19.0}{\scriptsize\,$\pm$\,0.1} \\
\midrule
\multirow{6}{*}{Llama-3.2-1B} & \multirow{2}{*}{SFT} & w/o CoT & \best{16.1} &
23.9{\scriptsize\,$\pm$\,0.2} & 41.7{\scriptsize\,$\pm$\,0.3} & 5.3{\scriptsize\,$\pm$\,0.1} & 9.5{\scriptsize\,$\pm$\,0.1} \\
 &  & w/ CoT & 420.3 &
 \underline{49.1}{\scriptsize\,$\pm$\,0.4} & \underline{64.7}{\scriptsize\,$\pm$\,0.3} & \underline{11.2}{\scriptsize\,$\pm$\,0.1} & \underline{15.3}{\scriptsize\,$\pm$\,0.1} \\
\noalign{\vskip 2pt}
\cdashline{2-8}
\noalign{\vskip 2pt}
 & \multirow{2}{*}{Coconut} & w/o Stage~0 & \multirow{2}{*}{108.3} &
 30.1{\scriptsize\,$\pm$\,0.3} & 46.1{\scriptsize\,$\pm$\,0.3} & 6.8{\scriptsize\,$\pm$\,0.1} & 11.0{\scriptsize\,$\pm$\,0.1} \\
 &  & w/ Stage~0 &  &
 36.9{\scriptsize\,$\pm$\,0.2} & 51.1{\scriptsize\,$\pm$\,0.2} & 8.5{\scriptsize\,$\pm$\,0.0} & 12.2{\scriptsize\,$\pm$\,0.0} \\
\noalign{\vskip 2pt}
\cdashline{2-8}
\noalign{\vskip 2pt}
 & \cellcolor{OursMainYellow} & \cellcolor{OursMainYellow}Final block &  \cellcolor{OursMainYellow} &
 \cellcolor{OursMainYellow}42.1{\scriptsize\,$\pm$\,0.2} & \cellcolor{OursMainYellow}56.1{\scriptsize\,$\pm$\,0.3} & \cellcolor{OursMainYellow}10.5{\scriptsize\,$\pm$\,0.0} & \cellcolor{OursMainYellow}13.8{\scriptsize\,$\pm$\,0.0} \\
 & \cellcolor{OursSoftYellow}\multirow{-2}{*}{\method{} (ours)} & \cellcolor{OursSoftYellow}Any block & \cellcolor{OursSoftYellow}\multirow{-2}{*}{\best{16.1}} &
 \cellcolor{OursSoftYellow}\best{51.4}{\scriptsize\,$\pm$\,0.2} & \cellcolor{OursSoftYellow}\best{76.8}{\scriptsize\,$\pm$\,0.1} & \cellcolor{OursSoftYellow}\best{13.0}{\scriptsize\,$\pm$\,0.0} & \cellcolor{OursSoftYellow}\best{19.6}{\scriptsize\,$\pm$\,0.1} \\
\midrule
\multirow{6}{*}{Llama-3.2-3B} & \multirow{2}{*}{SFT} & w/o CoT & \best{27.9} &
36.2{\scriptsize\,$\pm$\,0.2} & 45.9{\scriptsize\,$\pm$\,0.2} & 8.5{\scriptsize\,$\pm$\,0.1} & 10.8{\scriptsize\,$\pm$\,0.1} \\
 &  & w/ CoT & 754.4 &
 \best{66.9}{\scriptsize\,$\pm$\,0.2} & \best{78.3}{\scriptsize\,$\pm$\,0.4} & \best{19.0}{\scriptsize\,$\pm$\,0.1} & \best{24.4}{\scriptsize\,$\pm$\,0.3} \\
\noalign{\vskip 2pt}
\cdashline{2-8}
\noalign{\vskip 2pt}
 & \multirow{2}{*}{Coconut} & w/o Stage~0 & \multirow{2}{*}{188.8} &
 42.2{\scriptsize\,$\pm$\,0.2} & 56.8{\scriptsize\,$\pm$\,0.4} & 10.1{\scriptsize\,$\pm$\,0.0} & 13.3{\scriptsize\,$\pm$\,0.1} \\
 &  & w/ Stage~0 &  &
 41.3{\scriptsize\,$\pm$\,0.2} & 55.5{\scriptsize\,$\pm$\,0.5} & 10.2{\scriptsize\,$\pm$\,0.1} & 13.5{\scriptsize\,$\pm$\,0.1} \\
\noalign{\vskip 2pt}
\cdashline{2-8}
\noalign{\vskip 2pt}
 & \cellcolor{OursMainYellow} & \cellcolor{OursMainYellow}Final block & \cellcolor{OursMainYellow} &
 \cellcolor{OursMainYellow}48.8{\scriptsize\,$\pm$\,0.2} & \cellcolor{OursMainYellow}58.8{\scriptsize\,$\pm$\,0.2} & \cellcolor{OursMainYellow}12.0{\scriptsize\,$\pm$\,0.0} & \cellcolor{OursMainYellow}14.1{\scriptsize\,$\pm$\,0.0} \\
 & \cellcolor{OursSoftYellow}\multirow{-2}{*}{\method{} (ours)} & \cellcolor{OursSoftYellow}Any block & \cellcolor{OursSoftYellow}\multirow{-2}{*}{\best{27.9}} &
 \cellcolor{OursSoftYellow}\underline{57.3}{\scriptsize\,$\pm$\,0.1} & \cellcolor{OursSoftYellow}\underline{71.8}{\scriptsize\,$\pm$\,0.2} & \cellcolor{OursSoftYellow}\underline{13.8}{\scriptsize\,$\pm$\,0.1} & \cellcolor{OursSoftYellow}\underline{18.2}{\scriptsize\,$\pm$\,0.0} \\
\bottomrule
\end{tabular}
\vspace{-10pt}
\label{tab:appendix_main_results}
\end{table*}

\begin{table*}[b]
\centering
\small
\setlength{\tabcolsep}{8.5pt}
\renewcommand{\arraystretch}{1.2}
\caption{Linear probe separability and answer-selection accuracy by memory block. Values are reported in percent, and standard errors are percentage points computed over 16 split repeats.}
\label{tab:appendix_probe_separability}
\begin{tabular}{lcccccccc}
\toprule
 \multirow{2.3}{*}{Metric} & \multicolumn{5}{c}{Memory block} & 
 \multirow{2.3}{*}{\makecell{Probe-based\\Answer Selection}} \\
 \cmidrule(lr){2-6}
 & 1 & 2 & 4 & 6  & 8 & \\
\midrule
AUROC ($\uparrow$) & 
84.8{\scriptsize\,$\pm$\,0.1} & 
85.0{\scriptsize\,$\pm$\,0.1} & 
84.2{\scriptsize\,$\pm$\,0.1} & 
83.6{\scriptsize\,$\pm$\,0.1} & 
84.5{\scriptsize\,$\pm$\,0.1} & 
\textbf{86.0}{\scriptsize\,$\pm$\,0.1} \\
AUPRC ($\uparrow$) & 
80.7{\scriptsize\,$\pm$\,0.2} & 
82.3{\scriptsize\,$\pm$\,0.2} & 
82.0{\scriptsize\,$\pm$\,0.2} & 
81.6{\scriptsize\,$\pm$\,0.2} & 
81.9{\scriptsize\,$\pm$\,0.2} & 
\textbf{83.3}{\scriptsize\,$\pm$\,0.2} \\
\noalign{\vskip 1pt}
\cdashline{1-7}
\noalign{\vskip 1pt}
Accuracy ($\uparrow$) &  & & & & & \cellcolor{OursMainYellow}\textbf{90.0}{\scriptsize\,$\pm$\,0.2} \\
\bottomrule
\end{tabular}
\end{table*}

\paragraph{Probe-based answer selection.} We next ask whether the memory-block representations analyzed in \Cref{fig:appendix:pca_plots} contain enough information to identify the correct readouts.
For each memory block, we train a lightweight linear probe on the corresponding memory-block representations from 256 held-out GSM8K samples to predict whether the readout after that memory block is correct.
For an unseen evaluation sample, we group memory blocks whose generated answers are equivalent and combine the calibrated probe probabilities within each group, assigning higher confidence to answers supported by multiple blocks or by a highly confident block.
We then select the answer group with the highest combined confidence.
In \Cref{tab:appendix_probe_separability}, we report AUROC and AUPRC for these selected-answer confidence scores across 16 held-out splits, showing that correctness is highly predictable from the memory-block representations.
Conditioned on the recoverable subset where at least one memory block produces a correct answer, this selection procedure chooses a correct answer 90\% of the time.
This suggests that much of the gap between final-block and any-block accuracy can be closed with simple linear probes.
These results provide a simple proof of concept that correctness information is accessible from the memory-block representations, while more involved selection mechanisms are left to future work.

\begin{table*}[t]
\centering
\small
\setlength{\tabcolsep}{11pt}
\renewcommand{\arraystretch}{1.2}
\caption{\textbf{Official numbers.} Greedy accuracy on GSM8K and GSM-Hard, taking the official numbers from prior work. $\, ^{*}$official numbers from \cite{Jiang:25}, $\, ^{\dag}$official numbers from \cite{Hao:25}. }
\begin{tabular}{lcccccc}
\toprule
\multirow{2}{*}{Model}
& \multicolumn{3}{c}{GSM8K (ID)}
& \multicolumn{3}{c}{GSM-Hard (OOD)} \\
\cmidrule(lr){2-4}
\cmidrule(lr){5-7}
& DART & Coconut & \cellcolor{OursMainYellow}RiM (ours)
& DART & Coconut & \cellcolor{OursMainYellow}RiM (ours) \\
\midrule
GPT-2
& 24.7$^{*}$ & \underline{34.1}$^{\dag}$ & \cellcolor{OursMainYellow}\textbf{35.5}
& -- & -- & \cellcolor{OursMainYellow}8.4 \\

Llama-3.2-1B
& 42.6$^{*}$ & \textbf{50.6}$^{*}$ & \cellcolor{OursMainYellow}\underline{43.1}
& 10.9$^{*}$ & \textbf{11.2}$^{*}$ & \cellcolor{OursMainYellow}\textbf{11.2} \\

Llama-3.2-3B
& \underline{46.6}$^{*}$ & 43.6$^{\dag}$ & \cellcolor{OursMainYellow}\textbf{49.4}
& -- & -- & \cellcolor{OursMainYellow}12.1 \\
\bottomrule
\end{tabular}
\vspace{-4pt}
\label{appendix:tab:official_comparison}
\end{table*}

\begin{figure}[b]
    \centering

    \begin{subfigure}[t]{0.49\linewidth}
        \centering
        \includegraphics[width=\linewidth, trim=0pt 0pt 0pt 0pt, clip]{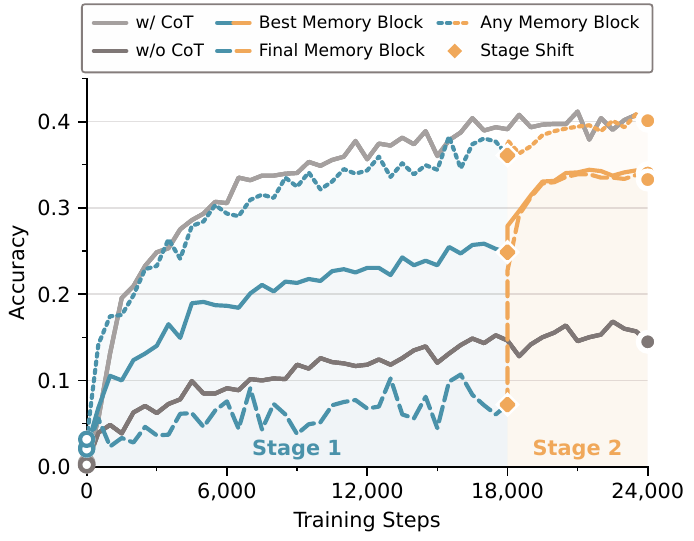}
        \caption{GPT-2}
        \label{fig:appendix:acc_sft_gpt2}
    \end{subfigure}
    \hfill
    \begin{subfigure}[t]{0.49\linewidth}
        \centering
        \includegraphics[width=\linewidth, trim=0pt 0pt 0pt 0pt, clip]{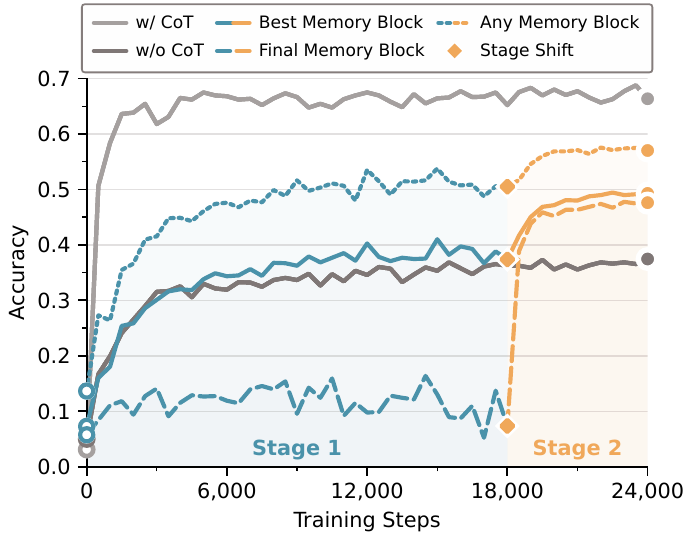}
        \caption{Llama-3.2-3B}
        \label{fig:appendix:acc_sft_llama3_3b}
    \end{subfigure}

    \caption{\textbf{GSM8K training curves.}
    Greedy accuracy on GSM8K test questions over training, comparing \method{} to SFT when training GPT-2 and Llama-3.2-3B on GSM8K-Aug.}
    \label{fig:appendix:acc_comparison}
\end{figure}

\paragraph{Comparison with Official Numbers.}
Official results from prior latent reasoning baselines are not fully controlled comparisons, as methods differ in their training setups and evaluation protocols. We therefore treat \Cref{appendix:tab:official_comparison} as a contextual comparison to the literature rather than as the primary evidence for our method. For completeness, we report \method{} under the same convention used for the official numbers. Instead of using the held-out cross-validation protocol from \Cref{tab:main_results}, we select the best checkpoint on the full evaluation set, allowing for a more direct comparison to the official numbers.

Even under this convention, the comparison remains conservative with respect to training cost. Under the Coconut method, GPT-2 is trained for 18 epochs \citep{Hao:25}, Llama-3.2-1B for 10 epochs \citep{Jiang:25}, and Llama-3.2-3B for 8 epochs \citep{Hao:25}. Moreover, the models are fine-tuned rather than trained with parameter-efficient LoRA adapters, which makes training substantially more expensive. Relative to our setup of 8 epochs with LoRA adapters, the number of training epochs alone corresponds to roughly 2.25$\times$ higher training cost for GPT-2 and 1.25$\times$ higher training cost for Llama-3.2-1B, before accounting for the additional cost of full-model fine-tuning.

The comparison to DART is even more conservative. Under the DART method, Llama-3.2-1B and Llama-3.2-3B are trained for 10 epochs each, and GPT-2 is trained for 40 epochs \citep{Jiang:25}. In addition, DART uses two separate training pathways that require two forward passes per sample and optimizes three loss terms simultaneously. Relative to our single-forward-pass training for 8 epochs, this corresponds to roughly 2.5$\times$ higher training cost for the Llama models and 10$\times$ higher training cost for GPT-2. 

Despite this larger training budget, \method{} outperforms the reported DART results in all comparable settings. More broadly, our method is best or tied for best in three of the four settings for which official DART or Coconut numbers are available. Thus, while differences in training setup preclude a strictly controlled evaluation, this official-number comparison shows that \method{} is highly competitive with substantially more involved latent reasoning methods.

\paragraph{Training curves across model scales.}
\Cref{fig:appendix:acc_comparison} complements the main training curves in \Cref{fig:acc_comp}, comparing \method{} with SFT using GPT-2 and Llama-3.2-3B.
The any-block accuracy rises substantially above direct-answer SFT, indicating that correct answers appear at some memory depths before the model is explicitly trained to use a fixed final block.
After the stage switch, Stage~2 rapidly transfers this latent computation into the deployable final-block readout, which improves over direct-answer SFT while retaining fixed answer-only inference latency.
The gap to explicit CoT remains larger for Llama-3.2-3B, but this baseline also spends additional decoding steps in language space and is therefore not latency-matched to \method{}.

\begin{figure}[t]
    \centering
    \includegraphics[width=\linewidth, trim = 0.0cm 0.0cm 0.0cm 0.0cm, clip]{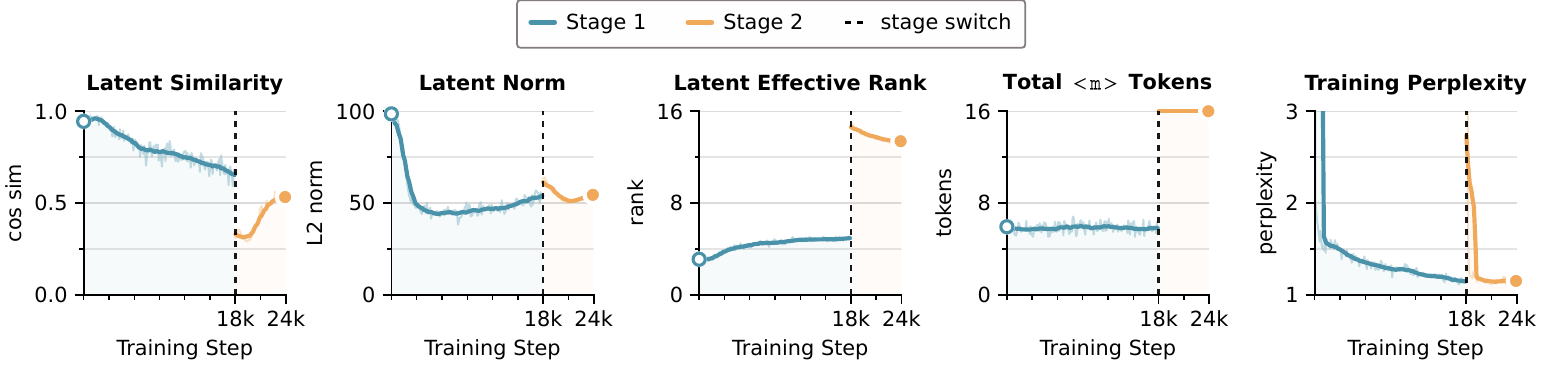}
    \caption{\textbf{Training diagnostics for \method{}.} From left to right: average within-block latent cosine similarity, average latent-state norm, latent effective rank, the active \texttt{<m>} token budget, and student training perplexity. The dashed line marks the transition from Stage~1 to Stage~2.}
    \label{fig:training_metrics}
\end{figure}

\paragraph{Training metrics.}
The geometric metrics in \Cref{fig:training_metrics} are computed from the final hidden layer of the latent tokens every 100 optimizer steps and averaged over distributed workers. We measure within-block latent similarity as the average pairwise cosine similarity between latent-token hidden states within the same contiguous memory block. This diagnostic captures whether tokens inside a memory block collapse to a common state or specialize into distinct directions. We also track the latent norm, computed as the average Euclidean norm of the latent-token hidden states. To measure the dimensionality of the latent workspace, we compute the effective rank from the singular-value spectrum of the matrix containing all latent-token states for each sample. Higher values indicate that the latent-token subspace uses more independent directions. We additionally log adjacent latent-state similarity, defined as the cosine similarity between consecutive latent tokens, as a smoothness diagnostic. Finally, student perplexity is the exponentiated token-averaged student NLL on the supervised training targets.

\begin{wraptable}{r}{0.44\textwidth}
\centering
\vspace{-13.4pt}
\setlength{\tabcolsep}{2pt}
\renewcommand{\arraystretch}{1.2}
\caption{\textbf{Inference latency.} Average tokens generated and wall-clock time in milliseconds (ms) per GSM8K question with Llama-3.2-1B, with standard errors across 4 runs.}
\label{tab:gsm8k-runtime-latency}
\begin{tabular}{lrrr}
\toprule
Method & Tokens & ms & $\Delta$ ms \\
\midrule
SFT w/o CoT  & 3.1  & 126.0{\scriptsize\,$\pm$\,0.5} & --- \\
\arrayrulecolor{gray!50}
\hdashline
\arrayrulecolor{black}
SFT w/ CoT  & 36.7 & 1108.7{\scriptsize\,$\pm$\,3.0} & {+982.7} \\
Coconut  & 3.1 & 304.7{\scriptsize\,$\pm$\,0.9} & {+178.7} \\
\cellcolor{OursMainYellow}RiM (ours) & \cellcolor{OursMainYellow}3.1 & \cellcolor{OursMainYellow}\best{126.5}{\scriptsize\,$\pm$\,0.5} & \cellcolor{OursMainYellow}{\best{+0.5}} \\
\bottomrule
\end{tabular}
\end{wraptable}

\paragraph{Empirical latency.}
\Cref{tab:gsm8k-runtime-latency} additionally reports the latency for generating the full answer, complementing the time to first token (TTFT).
The table records generated answer length and wall-clock time per GSM8K test question.
\method{} has essentially the same measured latency as SFT w/o CoT as the memory blocks are fixed input tokens processed in a single forward pass.
By contrast, SFT w/ CoT and Coconut are substantially slower because they require autoregressive generation of intermediate text tokens or CTs.
This makes the inference-cost comparison explicit and shows that \method{} preserves direct-answer inference speed.

\newpage
\newpage

\end{document}